\documentclass[lettersize,journal]{IEEEtran}
\usepackage{amsmath,amsfonts}
\usepackage{algorithmic}
\usepackage{algorithm}
\usepackage{array}
\usepackage[caption=false,font=normalsize,labelfont=sf,textfont=sf]{subfig}
\usepackage[colorlinks,linkcolor=red,citecolor=blue]{hyperref}
\usepackage{textcomp}
\usepackage{stfloats}
\usepackage{url}
\usepackage{verbatim}
\usepackage{graphicx}
\usepackage{cite}
\usepackage{bm}
\usepackage{fontawesome5}
\usepackage{soul}
\usepackage{color}
\usepackage{xcolor}
\usepackage{booktabs}
\usepackage{listings}
\usepackage{multirow}
\usepackage{diagbox}
\usepackage{graphicx}
\usepackage{caption}
\usepackage{makecell}

\newcommand{\revise}[1]{\textcolor{black}{#1}}
\newcommand{\revisesec}[1]{\textcolor{black}{#1}}

\begin{document}

\title{HuGDiffusion: Generalizable Single-Image Human Rendering via 3D Gaussian Diffusion}

\author{Yingzhi Tang, Qijian Zhang, and Junhui Hou, \textit{Senior Member}, \textit{IEEE}
\thanks{This work was supported in part by the NSFC Excellent Young Scientists Fund 62422118, and in part by the Hong Kong Research Grants Council under Grants 11219324, 11202320, and 11219422. (\textit{Corresponding author: Junhui Hou})}
\thanks{Yingzhi Tang and Junhui Hou are with the Department of Computer Science, City University of Hong Kong, Hong Kong SAR. E-mail: yztang4-c@my.cityu.edu.hk; jh.hou@cityu.edu.hk.}
\thanks{Qijian Zhang is with the TiMi L1 Studio of Tencent Games, China. Email: keeganzhang@tencent.com.}
}

\markboth{Manuscript submitted to IEEE TVCG}
{Shell \MakeLowercase{\textit{et al.}}: A Sample Article Using IEEEtran.cls for IEEE Journals}

\maketitle

\begin{abstract}
    We present HuGDiffusion, a generalizable 3D Gaussian splatting (3DGS) learning pipeline to achieve novel view synthesis (NVS) of human characters from single-view input images. Existing approaches typically require monocular videos or calibrated multi-view images as inputs, whose applicability could be weakened in real-world scenarios with arbitrary and/or unknown camera poses. In this paper, we aim to generate the set of 3DGS attributes via a diffusion-based framework conditioned on human priors extracted from a single image. Specifically, we begin with carefully integrated human-centric feature extraction procedures to deduce informative conditioning signals. Based on our empirical observations that jointly learning the whole 3DGS attributes is challenging  to optimize, we design a multi-stage generation strategy to obtain different types of 3DGS attributes. To facilitate the training process, we investigate constructing proxy ground-truth 3D Gaussian attributes as high-quality attribute-level supervision signals. Through extensive experiments, our HuGDiffusion shows significant performance improvements over the state-of-the-art methods. Our code will be made publicly available at \url{https://github.com/haiantyz/HuGDiffusion.git}.
\end{abstract}

\begin{IEEEkeywords}
Human Digitization, Novel View Synthesis, Neural Rendering, 3D Gaussian Splatting, Diffusion Model
\end{IEEEkeywords}

\section{Introduction}

\IEEEPARstart{T}{hree}-dimensional (3D) human digitization from single-view images has gained significant attention due to its wide-ranging applications in game production, filmmaking, immersive telepresence, and augmented/virtual reality (AR/VR). Recent advancements have led to the development of numerous methods leveraging deep learning architectures to infer 3D human from 2D image observations \cite{tian2023recovering}.

In terms of different learning objectives, existing 3D human digitization frameworks can be broadly categorized into two groups. \textit{Reconstruction-oriented} approaches \cite{saito2019pifu, zhang2023global, ho2024sith, zhang2024sifu} focus on recovering accurate 3D surface geometries of human characters while capturing corresponding appearance details, explicitly producing textured mesh models. However, these reconstructed meshes often suffer from issues such as incorrect poses, degenerate limbs, and overly smoothed surfaces. In contrast, \textit{rendering-oriented} approaches \cite{weng2022humannerf, hu2024gauhuman} prioritize visual presentation by synthesizing novel views from observed images, emphasizing rendering quality over precise geometric accuracy.

For human NVS, the primary objective is achieving high visual fidelity in rendered 2D images, irrespective of the accuracy of the underlying 3D geometric shapes. Most existing methods rely on video frame sequences or calibrated multi-view images as inputs, limiting their applicability in real-world scenarios with sparse views and unknown camera poses. To address this limitation, SHERF~\cite{hu2023sherf} introduce a generalizable NeRF~\cite{mildenhall2020nerf}-based learning framework for single-view human NVS in a feed-forward manner. However, the differentiable volume rendering technique used in SHERF necessitates a large number of query points along ray directions to render specific pixels, significantly reducing the efficiency of both training and inference processes.

\begin{figure*}[t]
    \centering
    \includegraphics[width=\textwidth]{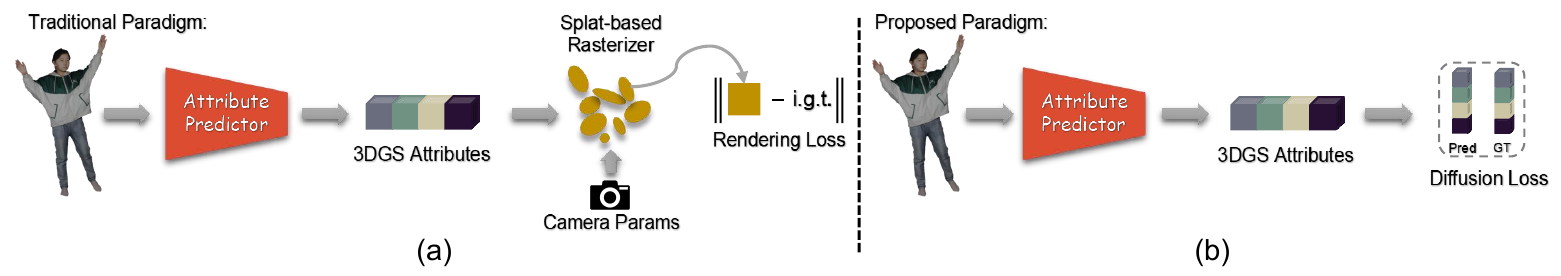}
    \caption{ (a) The traditional paradigm uses a splat-based rasterizer to supervise training with image signals.
(b) HugDiffusion trains a diffusion model with 3D Gaussian attributes as supervision signals.}
    \label{fig:teaser}
\end{figure*}

More recently, 3DGS \cite{kerbl20233d} rapidly evolve as a flexible and efficient neural rendering component. Inheriting the learning paradigms of generalizable NeRF frameworks \cite{yu2021pixelnerf,gao2022mps,hu2023sherf}, the latest works \cite{zheng2024gps,zou2024triplane} build generalizable 3DGS frameworks by training a parameterized neural model to generate the desired set of Gaussian attributes, which can be further rendered through splat rasterization. 
Technically, building a generalizable human 3DGS learning framework for single-view NVS faces several major aspects of challenges: 
\begin{enumerate}
    \item inferring complete 3D human appearance from a single 2D image is inherently a conditional generation task that requires the generation of invisible parts based solely on information from the visible view. Regression-based models lack the capability to predict accurate appearances in occluded areas, necessitating a more robust generation mechanism.;
    \item the prevailing training paradigm relies on regression models supervised by pixel-level signals (see Fig. \ref{fig:teaser}(a)). However, such signals are insufficient for training advanced generative models, such as diffusion models, highlighting the need for novel and more effective training signals;
    \item single-view images lack crucial features such as 3D human structure and unseen clothing details, making model training challenging. Human-centric conditions are necessary to bridge these gaps.
\end{enumerate}

To address these challenges, we propose a series of targeted solutions. We harness the diffusion mechanism to tackle this single-view human NVS as a conditional generation task (see Fig. \ref{fig:teaser}(b)). Specifically, we design a diffusion-based framework named HugDiffusion, which comprises two core modules. The initial module leverages a human reconstruction pipeline to generate point clouds, which are then initialized as 3D Gaussian positions. In the second module, we propose a conditional Gaussian attribute diffusion module to learn the data distribution of 3D Gaussian attributes, enabling the generation of realistic and plausible results.

To facilitate the training process of the diffusion model, we customize a two-stage workflow to pre-create 3D Gaussian attribute sets as attribute-level supervision signals. The workflow comprises a per-scene overfitting stage and a distribution unification stage. This workflow leverages point cloud transformers to learn structured 3D Gaussians and align attribute distributions across diverse scenes.

Additionally, human-centric conditions are constructed by integrating SMPL-Semantic features and Pixel-Aligned features based on the 3D Gaussian positions. These conditions provide essential context for accurate reconstruction.

In summary, our main technical contributions are:

\begin{itemize}
    \item we investigate a conditional diffusion framework for the generation of 3D Gaussian attribute sets, showing better generation capabilities than straightforward regression architectures;
    \item we propose a two-stage proxy ground-truth creation approach to achieve 3D Gaussian attribute sets as attribute-level supervision signals for the diffusion training process; and
    \item we develop a powerful human image feature extraction workflow, which effectively integrates human priors to extract more informative human-centric features to facilitate our conditional diffusion mechanism.
\end{itemize}

The remainder of this paper is organized as follows. Section \ref{sectionrelatedwork} provides a comprehensive review of the existing literature, including human reconstruction methods, novel view synthesis methods, and diffusion models. Section \ref{introduction} introduces our proposed two-stage proxy ground-truth construction process and HuGDiffusion in detail. In Section \ref{sectionexp}, we analyze the necessity of the proxy ground-truth construction process and conduct extensive experiments and ablation studies to validate the effectiveness of HuGDiffusion. Finally, Section \ref{sectionconclusion} concludes this paper.

\section{Related Work}
\label{sectionrelatedwork}
\subsection{Human Reconstruction}
The geometry of the human body serves as a fundamental basis for achieving accurate novel view synthesis. Various 3D representations, such as voxels, point clouds, meshes, and implicit functions, have been explored to represent human anatomy. Among these, implicit functions have emerged as a dominant approach for reconstructing geometric surfaces of the human body. PIFu and PIFuHD \cite{saito2019pifu, saito2020pifuhd} pioneered the use of implicit functions for monocular and multi-view human body reconstruction by predicting occupancy values of spatial points and extracting surfaces using the marching cubes algorithm. However, relying solely on image features without incorporating human body priors poses challenges in handling occluded regions.

To address this limitation, PAMIR and ICON \cite{zheng2021pamir, xiu2022icon} integrate the parametric human model SMPL to enable implicit functions to better comprehend the structure of the human body. Zhang et al. \cite{zhang2023global} further advanced the field by utilizing triplane representation and transformers to address issues such as loose clothing. SiFU \cite{zhang2024sifu} employs a text-to-image diffusion model to infer invisible details and produce realistic results. Despite these innovations, the reliance on SMPL introduces dependencies on accurate parameter estimation, as highlighted by Tang et al. \cite{tang2025human}, who demonstrated the challenges of refining SMPL parameters within implicit fields. Moreover, these methods often require querying a large number of spatial points to reconstruct 3D surfaces, which can be computationally intensive.

In contrast, HaP \cite{tang2025human}, a purely explicit method, represents the human body as a point cloud in 3D space, offering greater flexibility for modeling arbitrarily clothed human shapes with unconstrained topology. Due to its efficiency and adaptability, this work adopts HaP to generate 3D Gaussian positions.

\subsection{Novel View Synthesis}
View synthesis remains an open challenge in both academic research and industry applications. Recently, implicit neural radiance representation (NeRF) \cite{mildenhall2020nerf} has achieved remarkable success in generating high-quality novel views. Numerous NeRF-based methods have since been proposed to address various tasks, including single-view human NeRF \cite{hu2023sherf,weng2022humannerf} and multi-view human NeRF \cite{mu2023actorsnerf,gao2022mps}. By incorporating the parametric human model SMPL, NeRF-based methods \cite{jiang2022neuman,xu2021h} eliminate the need for ground-truth 3D geometry and enable the rendering of high-quality human views.
NeuralBody \cite{peng2021neural} introduced the use of NeRF for human novel view synthesis by learning structured latent codes for the canonical SMPL model across frames. HumanNerf \cite{weng2022humannerf} enhanced NeRF representations in canonical space by disentangling rigid skeleton motion from non-rigid clothing motion in monocular videos. Zhao et al. \cite{zhao2022humannerf} and Gao et al. \cite{gao2022mps} projected SMPL models onto multi-view images, combining image features with canonical SMPL features to create generalizable human NeRFs. SHERF \cite{hu2023sherf} introduced a hierarchical feature map for generalizable single-view human NeRF. These generalizable methods typically follow the training paradigm of PixelNeRF \cite{yu2021pixelnerf}. 

Recently, efforts have shifted toward generalizable 3D Gaussian splatting (3DGS) models \cite{liu2024fast,tang2024lgm,zou2024triplane,zheng2024gps,wang2024freesplat}, which offer faster training and inference compared to NeRF. Some video-based methods \cite{hu2024gauhuman,kocabas2024hugs} focus on learning 3D Gaussian attributes in canonical space for rapid rendering, but they typically rely on monocular or multi-view videos and lack generalizability across scenes. Zou et al. \cite{zou2024triplane} combined triplane and 3DGS representations, employing PIFu-like pixel-aligned features to train a generalizable 3DGS model. Zheng et al. \cite{zheng2024gps} proposed a framework for multi-view human novel view synthesis that integrates iterative depth estimation and Gaussian parameter regression.

Another emerging category of methods \cite{liu2023zero,yang2024magic,xue2024human} leverages the generative capabilities of large models \cite{rombach2022high} to predict unseen views. Tang et al. \cite{tang2024lgm} and Xue et al. \cite{xue2024human} combined diffusion-generated multi-view images with 3D information to generate 3D Gaussians. MagicMan \cite{he2024magicman} fine-tuned a stable diffusion model on human images and estimated SMPL models to synthesize novel human views. Similarly, our method utilizes a stable diffusion model \cite{brooks2023instructpix2pix} to infer unseen areas of a given human. 
\revise{Gen-3Diffusion \cite{xue2025gen} introduces a novel 3D-GS diffusion model for 3D reconstruction, which integrates large-scale priors from 2D multi-view diffusion models with efficient explicit 3D-GS representations through a sophisticated joint diffusion process to enhance 3D consistency. PSHuman \cite{li2025pshuman} recovers textured human meshes by generating multiview normal maps and color images.}

In this paper, a diffusion model is trained to predict 3D Gaussian distributions of the human body for novel view synthesis, employing attribute-level signals rather than pixel-level signals.

\begin{figure*}[t]
	\centering
	\includegraphics[width=6.8in]{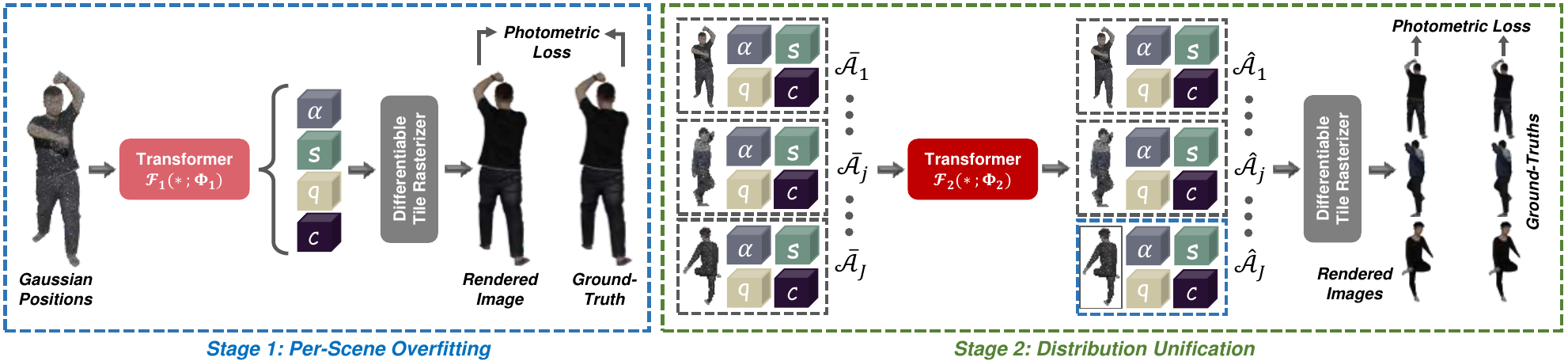}
	\caption{The two-stage workflow of creating proxy ground-truth Gaussian attributes.}
	\label{fig:pgt-creation}
\end{figure*}

\subsection{Diffusion Models} 
Diffusion models are particularly suited for addressing monocular human rendering tasks due to their strong capability to model 3D Gaussian attribute distributions from single-view images. These models have demonstrated remarkable success across various domains, including text-to-image generation \cite{zhang2023adding,rombach2022high}, super-resolution \cite{saharia2022image}, and low-light enhancement \cite{hou2024global}. Since 3D Gaussians can be conceptualized as point clouds with attributes, we briefly review advancements in point cloud diffusion models.

Luo et al. \cite{luo2021diffusion} pioneered the use of diffusion models for point cloud generation. Zhou et al. \cite{zhou20213d} extended this idea by developing a conditioned diffusion model based on partial point clouds. And diffusion models have also shown great potential in point cloud completion \cite{lyu2022a,cheng2025pvnet,wu20243d,wu2025unsupervised}.
Tang et al. \cite{tang2025human} adopted DDM \cite{ren2025ddm} at the refinement stage of the diffusion pipeline. Additionally, several methods \cite{vahdat2022lion,lyu2023controllable} have explored training point cloud diffusion models in latent spaces. However, these approaches are primarily focused on generating point cloud positions and are not directly applicable to the diffusion of 3D Gaussian attributes.
Among existing models, PC$^2$ \cite{melas2023pc2} is notable for being a point cloud diffusion model conditioned on single-view images, generating point clouds with color attributes. However, PC$^2$ does not integrate human body priors, which limits its ability to handle human-centric scenarios effectively.

Efforts have also been made to train 3D Gaussian splatting (3DGS) diffusion models. For instance, GaussianCube \cite{zhanggaussiancube} and L3DG \cite{roessle2024l3dg} construct voxel-based 3DGS representations and employ 3D U-Net architectures for training diffusion models. \revise{DiffGS \cite{zhou2024diffgs} utilizes a VAE to learn latent code with ground truth 3DGS attributes as supervision signals and then trains a latent diffusion model for various applications.} Despite their innovations, these methods primarily target generative tasks and often struggle to recover fine details from conditioning images \cite{zhanggaussiancube}. UVGS \cite{rai2025uvgs} attempts to diffuse the 3D Gaussian attributes by parametrizing them in the UV space \cite{zhang2023flattening,zeng2024dynamic}.

To address these limitations, we propose a tailored framework named HuGDiffusion, which leverages PointNet++ as its backbone. The framework incorporates meticulously designed human-centric conditioning mechanisms to enable effective diffusion of 3D Gaussian attributes, bridging the gap between generative capabilities and detailed reconstruction from single-view images.

\section{Proposed Method}
\label{introduction}
\subsection{Preliminary of 3DGS}

Different from implicit neural representation approaches \cite{mildenhall2020nerf,park2019deepsdf}, 3DGS~\cite{kerbl20233d} explicitly encodes a radiance field as an unordered set of Gaussian primitives denoted as $\mathcal{A} = \{ \mathbf{a}^{(n)} \}_{n=1}^{N}$. Each primitive is associated with a set of optimizable attributes: 

\begin{equation}
	\mathbf{a}^{(n)} = \{ \mathbf{p}^{(n)}, \alpha^{(n)}, \mathbf{s}^{(n)}, \mathbf{q}^{(n)},  \mathbf{c}^{(n)} \},
\end{equation}
including position $\mathbf{p}^{(n)} \in \mathbb{R}^3$, opacity value $\alpha^{(n)} \in \mathbb{R}$, scaling factor $\mathbf{s}^{(n)} \in \mathbb{R}^3$, rotation quaternion $\mathbf{q}^{(n)} \in \mathbb{R}^4$, and spherical harmonics (SH) coefficients $\mathbf{c}^{(n)} \in \mathbb{R}^d$. For an arbitrary viewpoint with camera parameters $\mathcal{V}$, a differentiable tile rasterizer $\mathcal{R}$ is applied to render the Gaussian attribute set $\mathcal{A}$ into the corresponding view image $\mathbf{I}_r$, which can be formulated as:
\begin{equation}
	\mathbf{I}_r = \mathcal{R}(\mathcal{A}; \mathcal{V}).
\end{equation}
For a set of observed $K$ multi-view images $\{ \mathbf{I}^{(k)} \}_{k=1}^{K}$ depicting a specific scene, together with their calibrated camera parameters $\{ \mathcal{V}^{(k)} \}_{k=1}^{K}$, the optimization process iteratively updates the Gaussian attributes by comparing the difference between rendered images and observed ground-truths, which can be formulated as:
\begin{equation}
	\mathbf{I}_r^{(k)} = \mathcal{R}(\mathcal{A}; \mathcal{V}^{(k)}),
	\quad
	\min_{\mathcal{A}} \sum\nolimits_{k=1}^K \ell_\mathrm{pmet}(\mathbf{I}_r^{(k)}, \mathbf{I}^{(k)}),
\end{equation}
\noindent where $\ell_\mathrm{pmet}(\cdot,\cdot)$ computes the pixel-wise photometric error within the image domain. After training, the resulting optimized Gaussian attribute set $\mathcal{A}$ serves as a high-accuracy neural representation of the target scene for real-time NVS. However, despite the fast inference speed of 3DGS, scene-specific overfitting still requires at least several minutes to complete.

\subsection{Training Paradigm of Generalizable Feed-Forward 3DGS}
In contrast to the conventional working mode of per-scene overfitting, many recent studies are devoted to constructing generalizable 3DGS frameworks \cite{liu2024fast,tang2024lgm,zou2024triplane,zheng2024gps,wang2024freesplat} via shifting the actual optimization target from the Gaussian attribute set to a separately parameterized learning model $\mathcal{M}(\ast; \Theta)$, where $\ast$ denotes network inputs and $\Theta$ denotes network parameters. Generally, we can summarize that all such approaches uniformly share the same training paradigm, where the learning model $\mathcal{M}(\cdot,\cdot)$ consumes its input to generate scene-specific Gaussian attributes at the output end. Through differentiable rendering $\mathcal{R}$, these approaches impose \textbf{\textit{pixel-level supervision}} with image signals, as formulated below:
\begin{equation}
	\min_{\Theta} \sum\nolimits_{k=1}^K \ell_\mathrm{pmet}(\mathcal{R}(\mathcal{M}(\ast; \Theta); \mathcal{V}^{(k)}), \mathbf{I}^{(k)}).
\end{equation}
Although such a training paradigm is reasonable and straightforward, its reliance on regression-based supervision limits the model's ability to accurately generate appearances in some occluded regions.

To overcome this limitation, we propose a paradigm shift by introducing \textbf{\textit{attribute-level supervision}}, enabling the effective training of diffusion-based models, which are well-known for their superior generative capabilities.
Under our targeted setting with single-view image $\mathbf{I}_s$ as input, the proposed training paradigm can be formulated as:
\begin{equation}
	\mathcal{A} = \mathcal{M}(\mathbf{I}_s; \Theta),
	\quad
	\min_{\Theta} \ell_\mathrm{setdiff}(\mathcal{A}, \hat{\mathcal{A}}),
\end{equation}
\noindent where $\ell_\mathrm{setdiff}(\cdot,~\cdot)$ measures the primitive difference between the predicted Gaussian attribute set $\mathcal{A}$ and the pre-created proxy ground-truth attribute set $\hat{\mathcal{A}}$.

Overall, our proposed single-view generalizable human 3DGS learning framework consists of two core processing phases: 1) \textit{creating proxy ground-truth Gaussian attributes as supervision signals}, and 2) \textit{training a conditional diffusion model for Gaussian attribute generation}, as introduced in the following Sections \ref{sec:proxy-gt-creation} and \ref{sec:gaus-attr-diffusion}.

\begin{figure*}[t]
	\centering
	\includegraphics[width=6.85in]{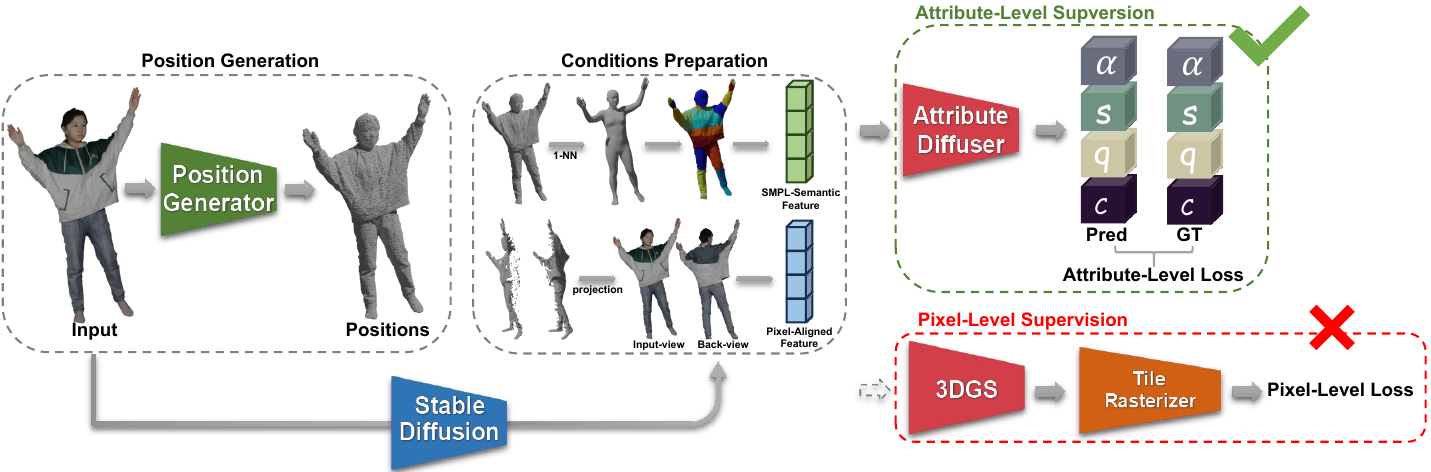}
	\caption{The framework of our HuGDiffusion.  HuGDiffusion predicts 3D Gaussian positions and a back-view image using a position generator and a stable diffusion module. It assigns SMPL semantic labels to points in 3D Gaussians, deducing an SMPL-semantic feature. The 3D Gaussians are decomposed for front- and back-view projection, achieving a pixel-aligned feature. Both features condition the 3D Gaussian attribute diffuser.}
	\label{HumanGSDiffframework}
\end{figure*}
\subsection{Creation of Attribute-Level Signals} \label{sec:proxy-gt-creation}

To facilitate attribute-level optimization, we need to pre-create a dataset of proxy ground-truth Gaussian attribute sets serving as the actual supervision signals for training $\mathcal{M}$. Formally, suppose that our raw training dataset is composed of $J$ different human captures each associated with multi-view image observations $\{ \mathbf{I}_j^{(k)} \}_{k=1}^{K}$ and camera parameters $\{ \mathcal{V}_j^{(k)} \}_{k=1}^{K}$, we aim to produce the corresponding proxy ground-truths $\{ \hat{\mathcal{A}} \}_{j=1}^{J}$ as:
\begin{equation}
	\hat{\mathcal{A}}_j = \{ \hat{\mathbf{a}}_j^{(n)} \}_{n=1}^{N} = \{ \hat{\mathbf{p}}_j^{(n)}, \hat{\alpha}_j^{(n)}, \hat{\mathbf{s}}_j^{(n)}, \hat{\mathbf{q}}_j^{(n)}, \hat{\mathbf{c}}_j^{(n)} \}.
\end{equation}

In fact, the most straightforward way of obtaining $\{ \hat{\mathcal{A}}_j \}_{j=1}^J$ is to separately overfit the vanilla 3DGS over each of the $J$ human captures and save the resulting Gaussian attributes. Unfortunately, owing to the inevitable randomness of gradient-based optimization and primitive manipulation, the overall distributions of the independently optimized Gaussian attribute sets are typically inconsistent. Even for the same scene, two different runs produce varying Gaussian attribute sets (e.g., primitive density and orders, attribute values), which results in a chaotic and challenging-to-learn solution space.

To obtain consistently distributed Gaussian attribute sets for shrinking the solution space, we particularly develop a two-stage proxy ground-truth creation workflow, as depicted in Fig.~\ref{fig:pgt-creation}. Given the task characteristics, we uniformly sample a dense 3D point cloud from the ground-truth human body surface to serve as the desired Gaussian positions $\{ \hat{\mathbf{p}}_j^{(n)} \}_{n=1}^N$. The other four types of attributes (i.e., opacities, scalings, rotations, SHs) are deduced from two sequential processing stages of what we call \textit{per-scene overfitting} and \textit{distribution unification}, as introduced below.

\subsubsection{Stage 1: Per-Scene Overfitting} This stage independently performs 3DGS overfitting on each of the $J$ human captures, but with one subtle difference from the vanilla optimization scheme. Specifically, instead of directly maintaining Gaussian attributes as learnable variables, we introduce a point cloud learning network $\mathcal{F}_1(\cdot;\Phi_1)$, which consumes $\{ \hat{\mathbf{p}}_j^{(n)} \}_{n=1}^N$ at the input end and outputs the rest types of Gaussian attributes, as formulated below: 
\begin{equation}
	\{ \bar{\alpha}_j^{(n)}, \bar{\mathbf{s}}_j^{(n)}, \bar{\mathbf{q}}_j^{(n)}, \bar{\mathbf{c}}_j^{(n)} \}_{n=1}^N =  \mathcal{F}_1( \{ \hat{\mathbf{p}}_j^{(n)} \}_{n=1}^N; \Phi_1 ).
\end{equation}
The purpose of moving Gaussian attributes to the output end of a neural network is to exploit the inherent smoothness tendency \cite{rahaman2019spectral} of the outputs of the neural network. Accordingly, for the $j$-th training sample, the per-scene optimization objective can be formulated as:
\begin{equation}
	\begin{split} & \bar{\mathcal{A}}_j = \{   \hat{\mathbf{p}}_j^{(n)}, \bar{\alpha}_j^{(n)}, \bar{\mathbf{s}}_j^{(n)}, \bar{\mathbf{q}}_j^{(n)}, \bar{\mathbf{c}}_j^{(n)} \}_{n=1}^N, \\ & \min \limits_{\Phi_1} \, \sum\nolimits_{k=1}^K \ell_\mathrm{pmet}(\mathcal{R}(\bar{\mathcal{A}}_j; \mathcal{V}_j^{(k)}); \mathbf{I}_j^{(k)}),
	\end{split}
\end{equation}
\noindent where $\ell_\mathrm{pmet}$ involves both $L_1$ and SSIM measurements. Additionally, auxiliary constraints are imposed over scaling and opacity attributes to suppress highly non-uniform distributions. The auxiliary constraints are as follows:	
\begin{equation}
	\Sigma_{n=1}^N\|\mathtt{radius}(\bar{\mathbf{s}}_j^{(n)})- \mathtt{kdist}(\hat{\mathbf{p}}_j^{(n)}) \|^2 + \Sigma_{n=1}^N \|  \bar{\alpha}_j^{(n)} -1 \|^2,
\end{equation}
where \( \mathtt{radius}(\cdot)\) and \(\mathtt{kdist}(\cdot)\) are the operations to get the radiuses of the Gaussians and the mean distances between the Gaussians and their neighbors. 

\subsubsection{Stage 2: Distribution Unification.} Though the preceding stage has preliminarily deduced a dataset of Gaussian attribute sets $\{ \bar{\mathcal{A}}_j \}_{j=1}^J$, the per-scene independent optimization can still lead to certain degrees of randomness and distribution inconsistency. \revise{To further align the distribution across different scenes}, in the second stage, we introduce another deep set architecture $\mathcal{F}_2(\cdot;\Phi_2)$ to overfit the whole $J$ training samples:
\begin{equation}
	\begin{split}
	 &\{ \hat{\mathcal{A}}_j \}_{j=1}^J =  \mathcal{F}_{2}( \{ \bar{\mathcal{A}}_j \}_{j=1}^J; \Phi_2 ), \\   \min \limits_{\Phi_2} \, & \sum\nolimits_{j=1}^J\sum\nolimits_{k=1}^K  \ell_\mathrm{pmet}(\mathcal{R}(\hat{\mathcal{A}}_j; \mathcal{V}_j^{(k)}); \mathbf{I}_j^{(k)}).
	\end{split}
\end{equation}
Since it is impractical to feed all $J$ training samples all at once, we adopt a batch-wise scheme with a certain number of training epochs, after which the resulting optimized $\{ \hat{\mathcal{A}} \}_{j=1}^{J}$ serves as our required proxy ground-truths.

\subsection{Gaussian Attribute Diffusion} \label{sec:gaus-attr-diffusion}

Having created a collection of proxy ground-truth Gaussian attributes $\{ \hat{\mathcal{A}}_j \}_{j=1}^J$ as supervision signals, we shift our attention to modeling the target distribution \(q(\mathcal{A}|\mathbf{I}_s)\) conditioned on the input single-view image $\mathbf{I}_s$ to predict the desired Gaussian attribute set $\mathcal{A}$. First, we separately predict Gaussian positions, which is essentially a 3D point cloud. Second, we treat the obtained point cloud as geometric priors and extract human-centric features and then feed them into a conditional diffusion pipeline for diffusing the rest types of Gaussian attributes.

\subsubsection{Generation of Gaussian Positions} In the training phase of the generalizable human 3DGS framework, we directly use the Gaussian positions $\{ \hat{\mathbf{p}}_j^{(n)} \}_{n=1}^N$ prepared in the proxy ground-truth creation process. In the inference phase, we need to estimate the Gaussian positions from the input image. In our implementation, we design a position generator with the rectification of the SMPL parametric human model.

The position generator begins with monocular depth estimation \cite{patni2024ecodepth} to generate from $\mathbf{I}_s$ the corresponding depth map, which is converted into a partial 3D point cloud. In parallel, we also estimate from $\mathbf{I}_s$ the corresponding SMPL model, whose pose is further rectified by the partial point cloud. Then, we feed the rectified SMPL model and the partial point cloud into a point cloud generation network to output the desired set of 3D Gaussian positions. To ensure the uniformity of point cloud, we perform point cloud upsampling and then apply farthest point sampling. In this way, we can stably obtain a set of accurate 3D Gaussian positions $\{ \mathbf{p}^{(n)} \}_{n=1}^N$ as geometric priors.

\subsubsection{Extraction of human-centric Features} To supplement more informative conditioning signals for the subsequent attribute diffusion, we further extract two aspects of human-centric features.

The first is \textit{pixel-aligned features} for providing visual appearance information. To achieve this, we project the 3D Gaussian positions onto the input image space. Then we utilize \cite{brooks2023instructpix2pix} to predict a back-view image \(\mathbf{I}_{\mathrm{back}}\) with respect to the view of $\mathbf{I}_s$ (\revise{We have fine-tuned the stable diffusion model on our training data}). The visible and invisible partitions of $\{ \mathbf{p}^{(n)} \}_{n=1}^N$ are respectively projected onto \(\mathbf{I}_s\) and \(\mathbf{I}_{\mathrm{back}}\), and the feature maps corresponding to \(\mathbf{I}_s\) and \(\mathbf{I}_{\mathrm{back}}\) are extracted via 2D CNNs. Finally, we concatenate the visible and invisible pixel-aligned features to form the pixel-aligned feature \(\bm{\beta}^{(n)}\).

The second is \textit{SMPL-semantic features}. \revise{To compensate for the lack of explicit spatial information in the unordered point cloud, we inject semantic labels defined on SMPL as structural priors. This enhances the model’s awareness of body configuration, thereby reducing noise and producing clearer boundaries between adjacent parts.}  We perform the nearest neighbor search to identify the nearest points of the 3D Gaussians on the SMPL surface. For each 3D Gaussian, we retrieve the nearest SMPL point index, distance, and semantic label, which are embedded into the latent space through MLPs. The resulting feature embeddings are concatenated to assign each point the SMPL-semantic feature \(\bm{\gamma}^{(n)}\).

\subsubsection{Conditional Diffusion} Having obtained Gaussian positions \({\mathbf{p}}^{(n)}\), pixel-aligned features \(\bm{\beta}^{(n)}\), and SMPL-semantic features \(\bm{\gamma}^{(n)}\), we perform condition diffusion to generate the rest attributes including $\alpha^{(n)}$, $\mathbf{s}^{(n)}$, $\mathbf{q}^{(n)}$, and $\mathbf{c}^{(n)}$. Empirically, we observe that simultaneously diffusing all these four types of attributes usually results in training collapse. We separate the diffusion of SH coefficients from the other three types of attributes.

For training the generation of SH coefficients, we design an attribute diffuser \( \mathtt{GSDIFF}_{\psi_1} \) to predict the noise at the given time step \( \mathbf{t}\) and use an $L_2$ loss for supervision:
\begin{equation}  
\begin{split}\epsilon^{(n)}  = &
 \mathtt{GSDIFF}_{\psi_1} (\tilde{\mathbf{c}}_t^{(n)}, {\mathbf{p}}^{(n)}, \bm{\beta}^{(n)},\bm{\gamma}^{(n)}, \mathbf{t} ), \\ & \quad
\min\limits_{\psi_1}  \mathbb{E}_{\epsilon \sim \mathcal{N}}\| \hat{\epsilon}^{(n)}-\epsilon^{(n)}\|^2,    
\end{split}
\end{equation}
where $\tilde{\mathbf{c}}_t^{(n)}$ denote SH coefficients with noise added, and \(\hat{\epsilon}^{(n)}\) is the ground-truth noise. For inference, we sample random SH coefficients \(\mathbf{c}^{(n)}_{\mathrm{T}}\) from the Gaussian distribution and iteratively remove noise to achieve \(\mathbf{c}^{(n)}_{\mathrm{0}}\). However, as demonstrated in PDR \cite{lyu2022a}, the inductive bias of the evidence lower bound (ELBO) is unclear in the 3D domain, resulting in \(\mathbf{c}^{(n)}_{\mathrm{0}}\) still containing noise, we further adopt an extra-step to remove the remaining noises. Also, we predict the other attributes, i.e., \( \alpha^{(n)}, \mathbf{s}^{(n)}, \mathbf{q}^{(n)}\) at this extra-step. 
\begin{equation}  
	\begin{split}
		  \{\epsilon^{(n)}, &\alpha^{(n)}, \mathbf{s}^{(n)}, \mathbf{q}^{(n)}\} = \mathtt{GSDIFF}_{\psi_2} (\mathbf{c}^{(n)}_{\mathrm{0}}, \mathbf{p}^{(n)}, \bm{\beta}^{(n)},\bm{\gamma}^{(n)}),  \\ & \quad  \quad \quad \quad \quad \mathbf{c}^{(n)} = \mathbf{c}_{\mathrm{0}}^{(n)} - \epsilon^{(n)}, \\   &\min\limits_{\psi_2} \| \mathbf{c}^{(n)}- \hat{\mathbf{c}}^{(n)}\|^2 +\| \alpha^{(n)}-\hat{\alpha}^{(n)}\|^2+ \\ &\quad \quad 
          \| \mathbf{s}^{(n)}-\hat{\mathbf{s}}^{(n)}\|^2+\| \mathbf{q}^{(n)}-\hat{\mathbf{q}}^{(n)}\|^2,    
	\end{split}
\end{equation}

Finally, we obtain a 3D Gaussian attribute set \(\{\mathbf{p}^{(n)}, \alpha^{(n)}, \mathbf{s}^{(n)}, \mathbf{q}^{(n)},\mathbf{c}^{(n)}\}\) of a scene when given a single-view image \(\mathbf{I}_s\), which can be used to render novel views of the human body. 

\begin{figure*}[t]
	\centering
	\includegraphics[width=6.8in]{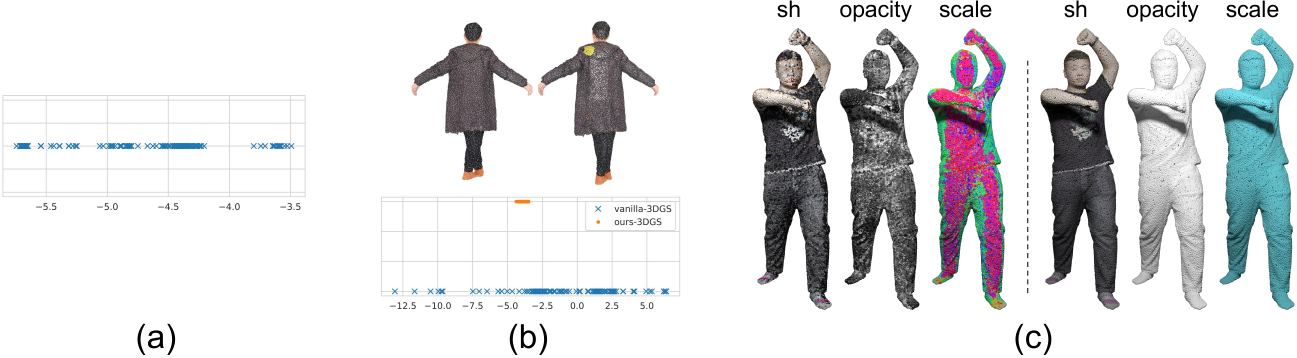}
	\caption{  (a) The large variation of vanilla-3DGS spherical harmonic values after 200 attempts on a scene. (b) Spherical harmonic values comparison of the local area (marked with \textcolor{yellow}{yellow}) between vanilla-3DGS and our proxy ground-truth 3D Gaussian attributes. (c) Visualization of spherical harmonic, opacity, and scale for vanilla-3DGS and our proxy ground-truth 3D Gaussian attributes.  \textcolor{red}{\faSearch} Zoom in for details.}
	\label{validatevanila3dgs}
\end{figure*}

\section{Experiments}
\label{sectionexp}
\subsection{Datasets and Implementation Details}
\label{reproduce}
\begin{table*}[t]
	\scriptsize
	\centering
	\renewcommand\arraystretch{1.25}
	\caption{\revise{Quantitative comparisons of different methods on Thuman, CityuHuman, 2K2K, and CustomHuman datasets. The best results are highlighted in \textbf{bold}. $\uparrow$: the higher the better. $\downarrow$: the lower the better.}}
	\resizebox{0.97\linewidth}{!}{\begin{tabular}{l|c|c|c|c|c|c|c|c|c|c|c|c} 
			\toprule
			\multicolumn{1}{c|}{\multirow{2}[0]{*}{\diagbox{Method}{Metric}}} & \multicolumn{3}{c|}{Thuman} & \multicolumn{3}{c|}{CityuHuman} & \multicolumn{3}{c|}{2K2K} & \multicolumn{3}{c}{CustomHuman}    \\
			& \multicolumn{1}{c}{PSNR$\uparrow$} & \multicolumn{1}{c}{SSIM$\uparrow$} & \multicolumn{1}{c|}{LPIPS$\downarrow$} & \multicolumn{1}{c}{PSNR$\uparrow$} & \multicolumn{1}{c}{SSIM$\uparrow$} & \multicolumn{1}{c|}{LPIPS$\downarrow$} & \multicolumn{1}{c}{PSNR$\uparrow$} & \multicolumn{1}{c}{SSIM$\uparrow$} & \multicolumn{1}{c|}{LPIPS$\downarrow$}   & \multicolumn{1}{c}{PSNR$\uparrow$} & \multicolumn{1}{c}{SSIM$\uparrow$} & \multicolumn{1}{c}{LPIPS$\downarrow$}\\
			\hline
			\hline
			GTA ~\cite{zhang2023global} & 25.78   & 0.919  & 0.085  & 27.41  & 0.923  & 0.075& 24.15  & 0.921  & 0.080 &28.86 &0.920&0.088 \\ 
			SiTH ~\cite{ho2024sith} & 25.36 & 0.919 & 0.083 & 29.21 &0.934 &0.067 & 24.30
			& 0.920 & 0.076 &26.47 &0.911&0.095 \\
			LGM ~\cite{tang2024lgm} & 25.13  & 0.915 & 0.096 & 29.78   & 0.941 & 0.074 &27.99 & 0.938&0.071 &31.91& 0.944 &0.077 \\ 
			SHERF ~\cite{hu2023sherf} & 26.57   & 0.927   & 0.081 & 30.13   & 0.942   & 0.067 &27.29&0.931&0.072&27.88&0.916&0.096\\ 
            SIFU \cite{zhang2024sifu} & 23.16
			& 0.904 & 0.102 & 26.46
			& 0.917 & 0.087& 24.30
			& 0.920 & 0.076& 29.62 & 0.928 & 0.092\\
			Human-3Diffusion~\cite{xue2024human} & 27.06   & 0.934   & 0.079 & 30.48   & 0.944   & 0.068 &29.05&0.942&0.062&33.75&0.952&0.067\\
            PSHuman~\cite{li2025pshuman} & 25.34  & 0.910   & 0.084 & 27.82   & 0.925   & 0.071 & 24.72 & 0.917 & 0.067 & 30.26 & 0.931 & 0.082\\
			\hline
            HuGDiffusion \textcolor{gray}{Neural} & 29.70  & 0.950   & 0.069 & 32.39   & 0.953   & 0.064 & 30.18 & 0.947 & 0.062 & 34.64 & 0.953 & 0.064 \\
			HuGDiffusion \textcolor{gray}{Joint} & \textbf{30.03} & \textbf{0.953}   & \textbf{0.065} & \textbf{32.47}  & \textbf{0.954} & \textbf{0.062}  & \textbf{30.64}  & \textbf{0.949} & \textbf{0.060} & \textbf{34.82}  & \textbf{0.958} & \textbf{0.055} \\ 
			\bottomrule
	\end{tabular}}
	\label{sotaresults}
\end{table*}

We utilized 480 human models from Thuman2 \cite{yu2021function4d} for the construction of proxy ground truth 3D Gaussian attributes and the training of the attribute diffusion model. We quantitatively evaluated HuGDiffusion on Thuman2 (20 humans), CityuHuman  (20 humans) \cite{tang2025human}, 2K2K (25 humans) \cite{han2023high} and CustomHuman (40 humans) \cite{ho2023learning}. \revise{The images are rendered with Blender in $512 \times 512$ resolution}. We adopted the peak signal-to-noise ratio (PSNR), structural similarity index (SSIM), and Learned Perceptual Image Patch Similarity (LPIPS) as evaluation metrics \revise{on the entire images}.

When constructing the proxy ground truth 3D Gaussian attributes, we rendered 360 views for each human, and we uniformly sampled 20000 points from each human surface, which was the initial 3D Gaussian position. In the first stage, we utilized a Point Transformer as the backbone to predict the 3D Gaussian attributes. We overfitted the Point Transformer for 4000 epochs for each human subject, using the Adam optimizer with a learning rate of 0.0002. In the second stage, we employed another Point Transformer as the backbone. the batch size was set to 4, the number of epochs to 1300, and we continued using the Adam optimizer with a learning rate of 0.0002. Other settings of 3DGS followed \cite{kerbl20233d}. 
To train HuGDiffusion, we designed the attribute diffusion model, with PointNet++ serving as the backbone\footnote{We found that the voxel size of the point transformer will seriously affect the performance. Hence, we did not use it as the backbone of  HuGDiffusion.}. ResNet18 with pre-trained weights was used to extract image features when preparing the pixel-aligned feature. During training, the batch size was set to 4, the number of epochs was set to 300, the optimizer was Adam, and the learning rate was 0.0002. 

\begin{table}[t]
\centering
	\renewcommand\arraystretch{1.25}
\caption{The results of different settings on Thuman. Results: Ground truth point clouds are used. \textcolor{gray}{\it Results: generated point clouds are used. }}
\label{validation0}
\resizebox{1.0\linewidth}{!}{\begin{tabular}{c|c|ccc}
\toprule
\multirow{2}{*}{} & \multirow{2}{*}{Pixel-Level} & \multicolumn{3}{c}{Attribute-Level}                                             \\ \cline{3-5} 
                  &                              & \multicolumn{1}{c|}{vanilla} & \multicolumn{1}{c|}{neural}       & joint        \\ \hline \hline
PSNR $\uparrow$             & 31.84 \textcolor{gray}{({\it 29.02})}                 & \multicolumn{1}{c|}{FAIL}    & \multicolumn{1}{c|}{32.99 \textcolor{gray}{({\it 29.53})}} & 33.22 \textcolor{gray}{({\it 29.71})} \\ 
SSIM   $\uparrow$           & 0.963 \textcolor{gray}{({\it 0.953})}                 & \multicolumn{1}{c|}{FAIL}    & \multicolumn{1}{c|}{0.968 \textcolor{gray}{({\it 0.949})}} & 0.971 \textcolor{gray}{({\it 0.951})} \\ 
LPIPS   $\downarrow$          & 0.056 \textcolor{gray}{({\it 0.070}) }                & \multicolumn{1}{c|}{FAIL}    & \multicolumn{1}{c|}{0.052 \textcolor{gray}{({\it 0.069})}} & 0.048 \textcolor{gray}{({\it 0.065})} \\ \bottomrule
\end{tabular}}
\end{table}

\begin{figure}
	\begin{center}
		\includegraphics[width=3.5in]{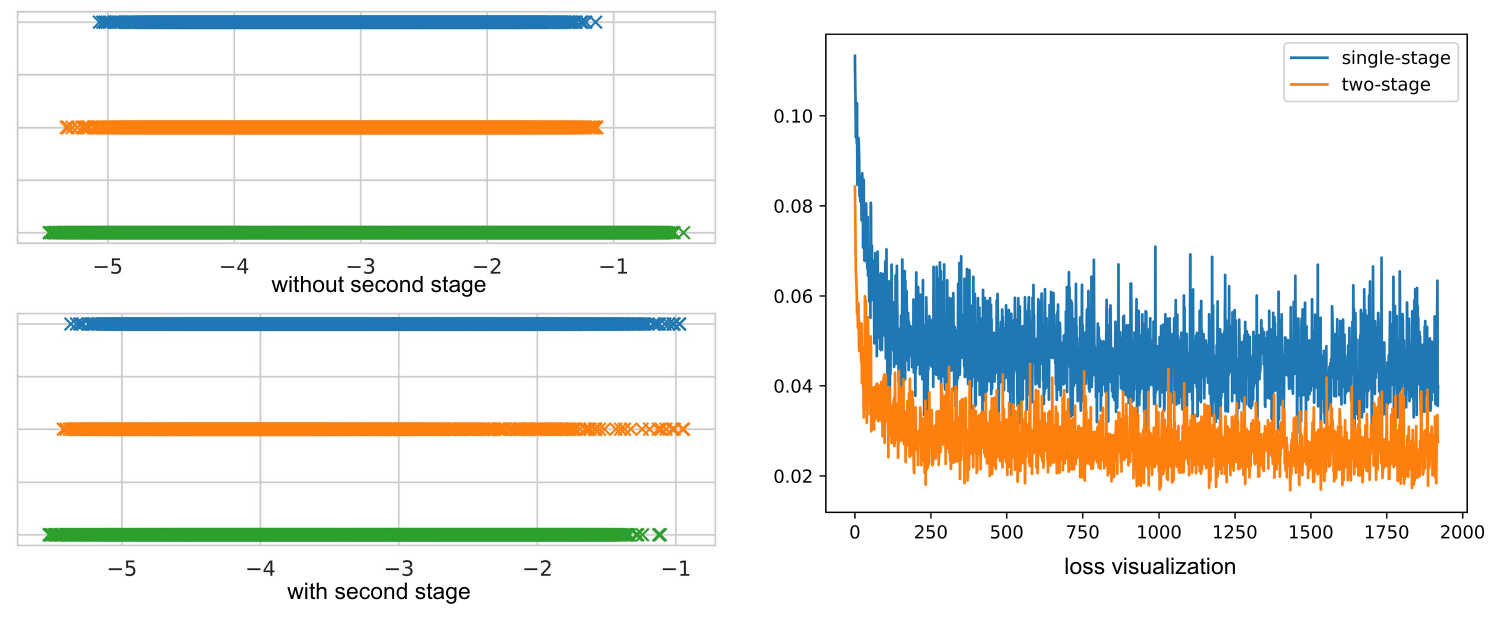}
	\end{center}
	\caption{The visualization of distributions and loss.  \textcolor{red}{\faSearch} Zoom in for details.}
	\label{wosecondstage}
\end{figure}
\begin{figure}[t]
	\begin{center}
		\includegraphics[width=3.2in]{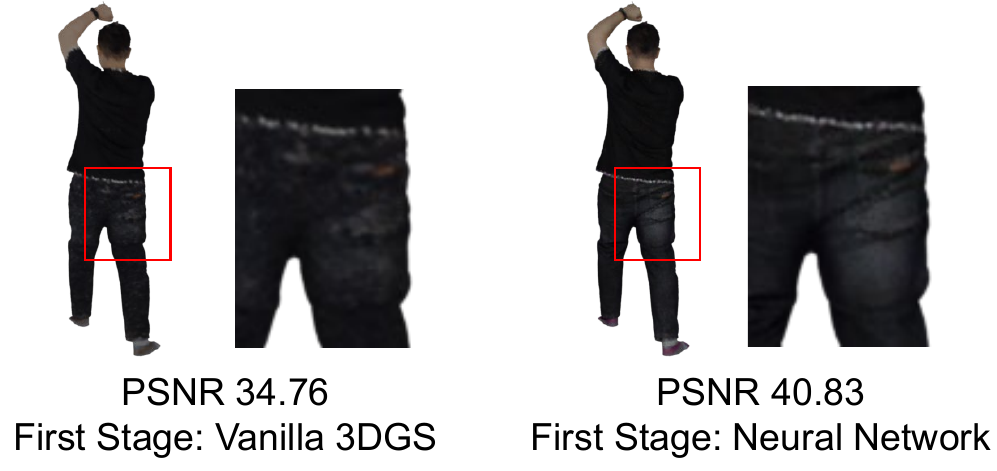}
	\end{center}
	\caption{The visual comparison with vanilla-3DGS and point transformer at the first stage.  \textcolor{red}{\faSearch} Zoom in for details.}
	\label{firststagevanila}
\end{figure}

\begin{figure}[t]
	\begin{center}
		\includegraphics[width=3.2in]{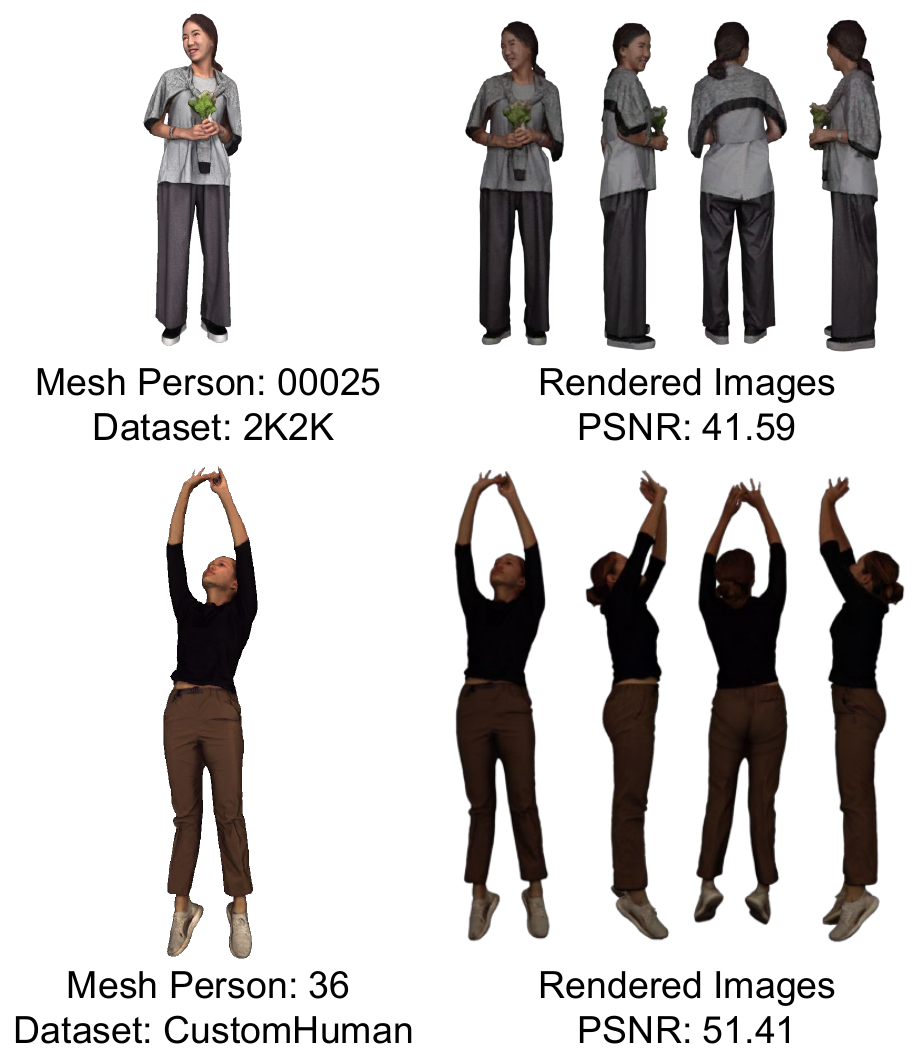}
	\end{center}
	\caption{The constructed results on 2K2K and CustomHuman datasets.  \textcolor{red}{\faSearch} Zoom in for details.}
	\label{generalizability2s}
\end{figure}
 \begin{figure*}[t]
	\centering
	\includegraphics[width=6.85in]{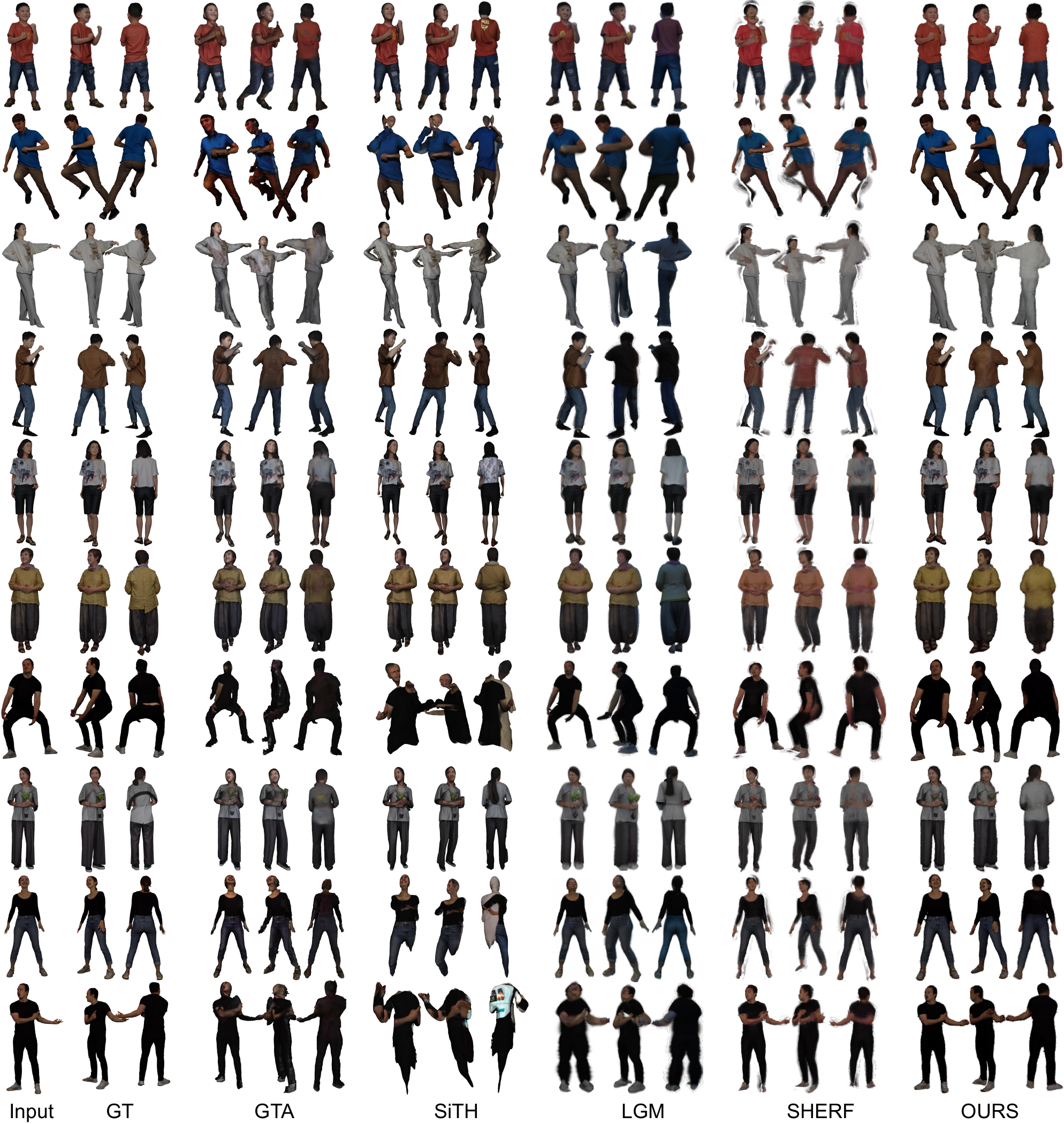}
	\caption{Visual comparisons of our method against GTA~\cite{zhang2023global}, SiTH~\cite{ho2024sith}, LGM~\cite{tang2024lgm}, and Sherf~\cite{hu2023sherf}. \textcolor{red}{\faSearch} Zoom in for details.}
	\label{comparesota}
\end{figure*}

 \begin{figure*}[t]
	\centering
	\includegraphics[width=6.85in]{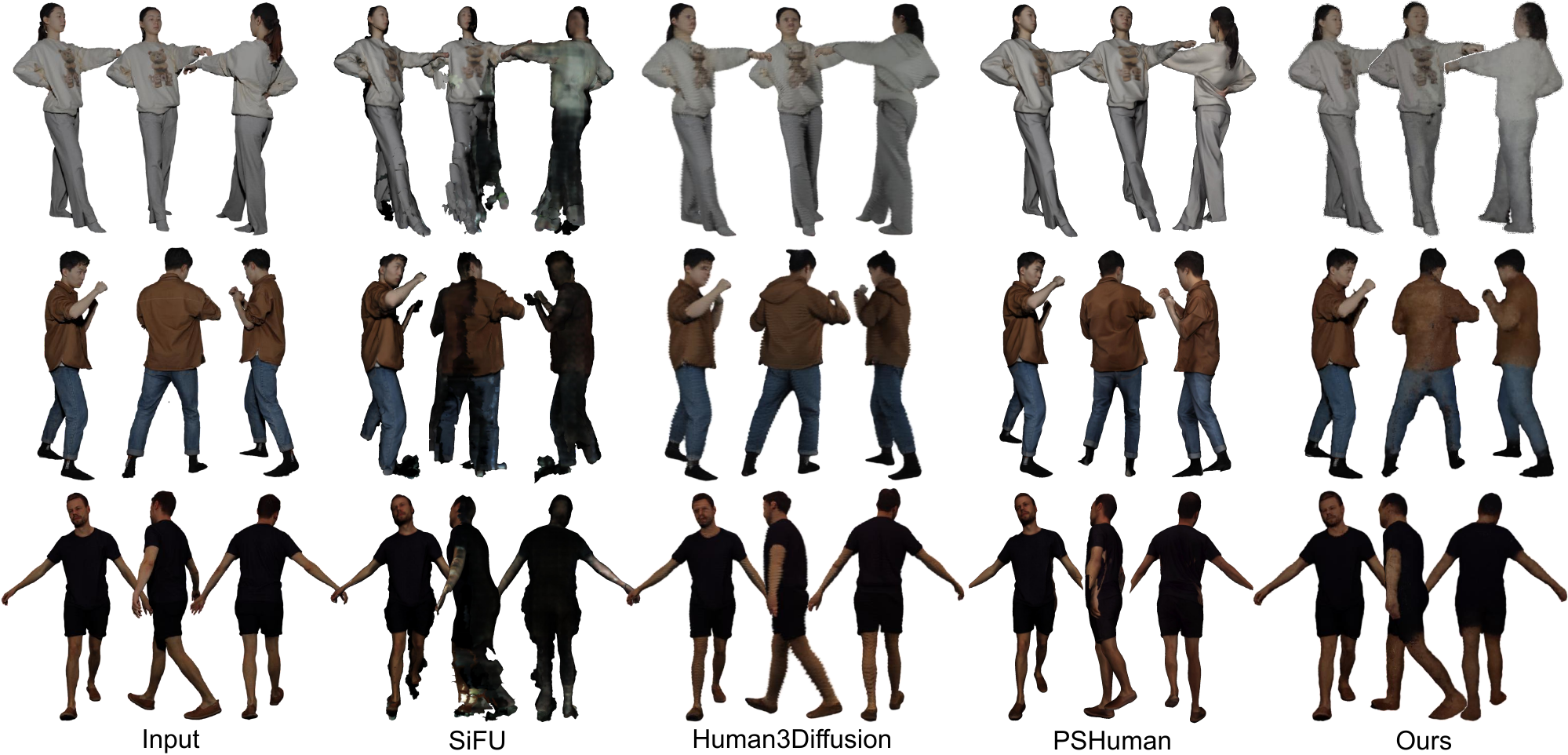}
	\caption{Visual comparisons of our method against SiFU~\cite{zhang2024sifu}, Human-3Diffusion~\cite{xue2024human}, and PSHuman~\cite{li2025pshuman}. \textcolor{red}{\faSearch} Zoom in for details.}
	\label{comparesota2}
\end{figure*}
\subsection{Analysis on 3D Gaussian Attribute Construction}

We first conducted experiments to demonstrate why point transformers are necessary for overfitting and why a two-stage workflow is required. \revise{The quantitative results are shown 
in Tab. \ref{validation0}. (Note that all experimental results in this table are trained in regression manner.)}

\textbf{Defect of Vanilla-3DGS.} The solution space of vanilla-3DGS is expansive, primarily due to directly optimizing numerical values and the strong interdependency of its attributes. To validate this, we designed three experiments. Firstly, we fixed all random seeds in \(\mathtt{python}\), \(\mathtt{numpy}\) and \(\mathtt{pytorch}\) and optimized the 3DGS model on a specific scene 200 times. Subsequently, we plotted the results from 200 optimizations, obtained by summing the spherical harmonic values of each optimization, as depicted in Fig. \ref{validatevanila3dgs}(\textcolor{red}{a}). The results exhibit significant variation. Secondly, we selected a local area (marked in \textcolor{yellow}{yellow}) on a human body expected to have the same color; however, as illustrated in Fig. \ref{validatevanila3dgs}(\textcolor{red}{b}), the spherical harmonic values ranged widely from -12.5 to 6. Lastly, we visualized the spherical harmonic, opacity, and scale attributes in Fig. \ref{validatevanila3dgs}(\textcolor{red}{c}) left column, revealing that vanilla-3DGS produced highly chaotic results. Although vanilla-3DGS can  better rendering results, its inherent randomness and lack of regularity make it impractical for learning (as illustrated in Tab. \ref{validation0}). Therefore, it is not suitable for constructing the proxy ground truth 3D Gaussian attributes.

\begin{figure}[t]
	\begin{center}
		\includegraphics[width=3.2in]{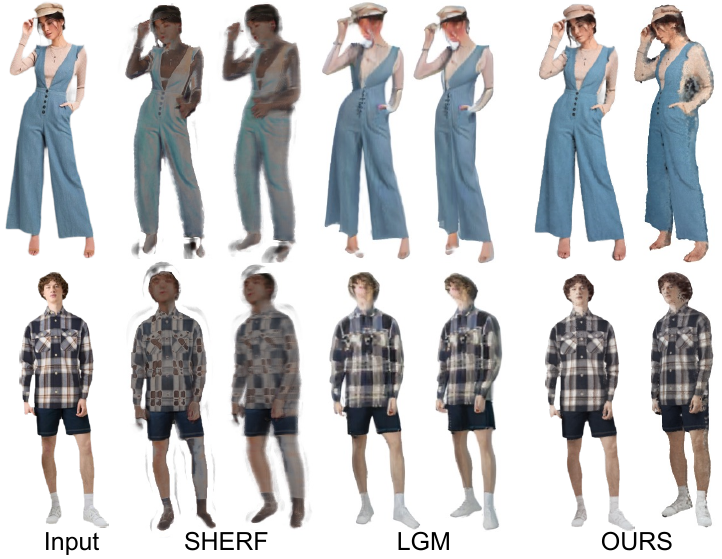}
	\end{center}
	\caption{The visual comparison with SHERF~\cite{hu2023sherf} and LGM~\cite{tang2024lgm} on the wild images.  \textcolor{red}{\faSearch} Zoom in for details.}
	\label{inthewildcompare}
\end{figure}

 \begin{figure*}[t]
	\centering
	\includegraphics[width=6.85in]{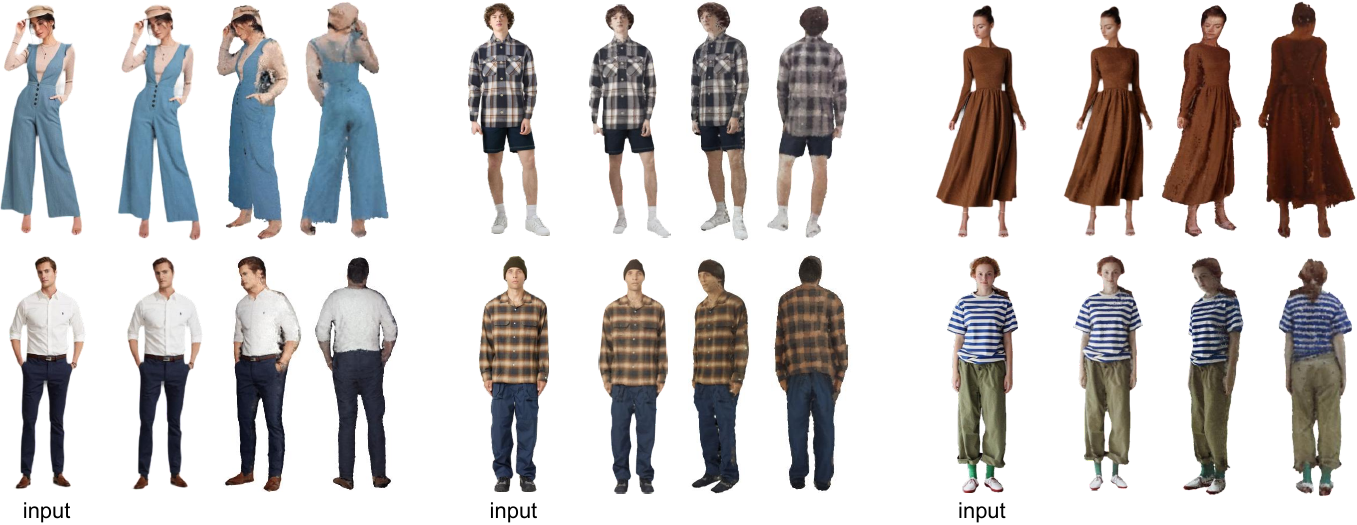}
	\caption{Rendering results of HuGDiffusion on the in-the-wild images. \textcolor{red}{\faSearch} Zoom in for details.}
	\label{inthewild1}
\end{figure*}

\begin{figure}[t]
	\begin{center}
		\includegraphics[width=3.2in]{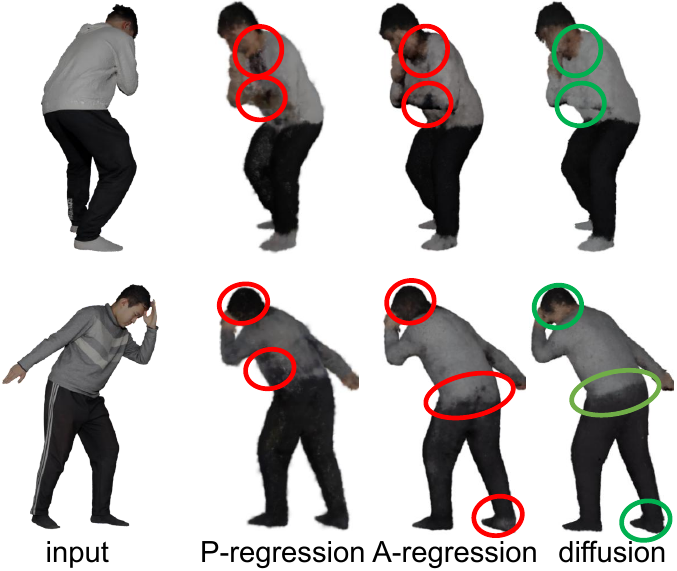}
	\end{center}
	\caption{Regression-based model v.s. diffusion-based model. P-regression: supervised with pixel-level signals. A-regression: supervised with attribute-level signals. \textcolor{red}{\faSearch} Zoom in for details.}
	\label{regressionisbad}
\end{figure}

\begin{figure}[t]
	\begin{center}
		\includegraphics[width=3.2in]{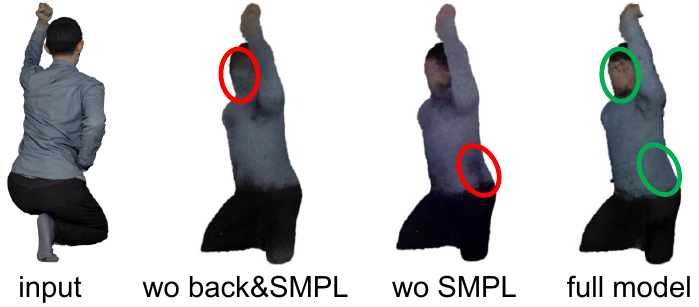}
	\end{center}
	\caption{Without occluded side image v.s. with occluded side image. \textcolor{red}{\faSearch} Zoom in for details.}
	\label{woback}
\end{figure}
\begin{figure}[t]
	\begin{center}
		\includegraphics[width=3.5in]{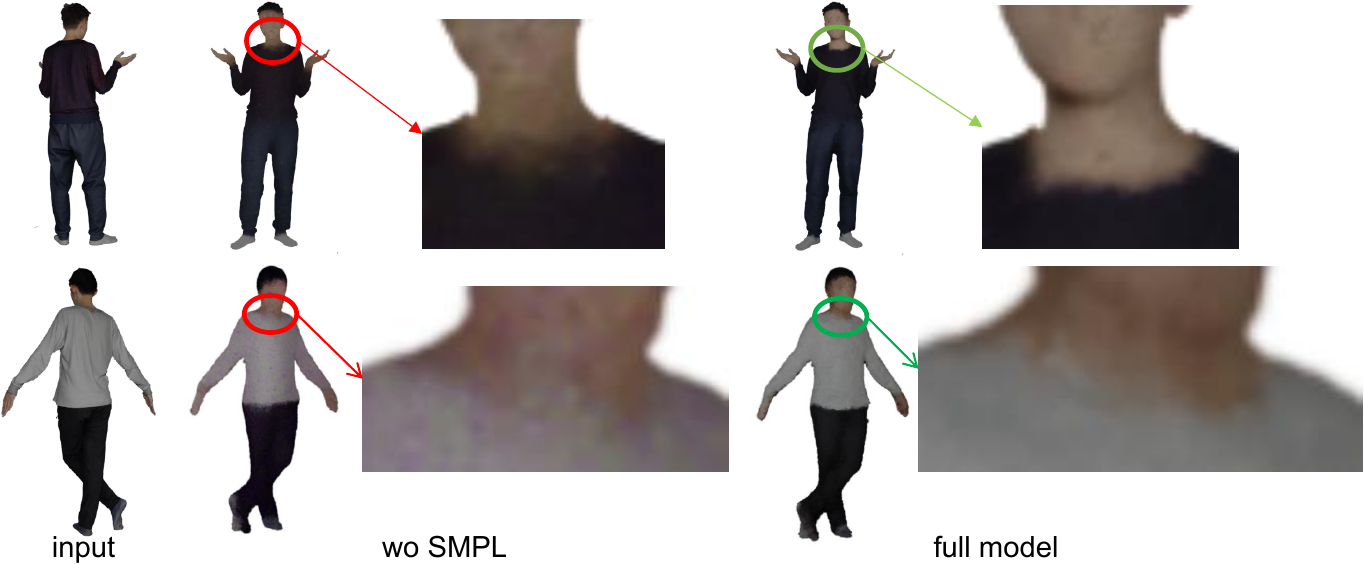}
	\end{center}
	\caption{Without SMPL semantic features v.s. with SMPL semantic features \textcolor{red}{\faSearch} Zoom in for details.}
	\label{wosmpl}
\end{figure}

\textbf{The Necessity of two-stage construction.}  We visualized the results of the per-scene overfitting stage in Fig. \ref{validatevanila3dgs} (\textcolor{red}{b}) and (\textcolor{red}{c}) right column. Our point transformer-based 3DGS significantly reduced the variation range of spherical harmonic values in the local area, and the visualizations of the spherical harmonic, opacity, and scale attributes appeared cleaner and more uniform.

However, as depicted in Fig. \ref{wosecondstage}, the 3D Gaussian attributes obtained from the first stage resulted in slow convergence during training, primarily due to the independent optimization of scenes, which caused distinct distributions across different scenes. After the second stage, the distribution among various scenes was better aligned, with the variances of the minimum and maximum values reduced from \textbf{0.0707} to \textbf{0.0329} and from \textbf{0.1114} to \textbf{0.1094}, respectively. As shown in Tab. \ref{validation0}, the two-stage construction (\textbf{joint}) achieves better performance than the single-stage construction (\textbf{neural}).

We also directly applied the vanilla-3DGS in the first stage. However, the high level of randomness in the vanilla-3DGS could not be sufficiently mitigated by relying solely on the point transformer in the second stage, as illustrated in Fig. \ref{firststagevanila}. This further demonstrates the necessity of using the point transformer in the first stage. Hence, a two-stage construction process is essential.

\textbf{Generalizability to Other Datasets.} To validate the generalizability of our two-stage construction process, we applied it to the 2K2K and CustomHuman datasets. As illustrated in Fig. \ref{generalizability2s}, the process consistently produced satisfactory results across these datasets.

\subsection{ Comparisons with State-of-the-Art Methods}
We compared our HuGDiffusion with state-of-the-art methods: GTA \cite{zhang2023global}, LGM \cite{tang2024lgm}, SiTH \cite{ho2024sith}, and SHERF \cite{hu2023sherf}, \revise{SiFU \cite{zhang2024sifu}, Human-3Diffusion \cite{xue2024human} and PSHuman \cite{li2025pshuman}}. The four used datasets were collected from individuals of different ages, genders, and races.  As reported in Tab. \ref{sotaresults}, HuGDiffusion achieves the best quantitative performance across all metrics on all datasets, underscoring the effectiveness and generalization capability of HuGDiffusion.

GTA and SiTH suffer from the grid resolution of the marching cube and produce broken reconstructed human bodies, resulting in low-fidelity novel views. Moreover, their results are usually in the wrong poses. However, as shown in Fig. \ref{comparesota}, our HuGDiffusion is capable of rendering fine-grained input view images while maintaining texture consistency across different view directions with correct poses. As shown in Fig. \ref{comparesota2}, the reconstructed body generated by SiFU \cite{zhang2024sifu} is fragmented and fails to recover the correct appearance, while PSHuman \cite{li2025pshuman} often suffers from pose misalignment and missing body parts. \revisesec{GTA, SiTH, and SiFU are implicit approaches without accurate ground-truth appearance supervision. They rely on nearest-neighbor color sampling during training, which can limit their ability to learn precise appearance details. In contrast, HuGDiffusion is trained with explicit ground-truth 3D Gaussian attributes, thus leading to superior performance in appearance reconstruction. PSHuman learns multi-view human images and it employs continuous remeshing \cite{palfinger2022continuous} to convert 2D normal maps into 3D meshes, however, such an optimization process often leads to overfitting, resulting in missing body parts and degraded appearance rendering from novel views. In contrast, HuGDiffusion learns a complete 3D point cloud structure, enabling more consistent geometry and superior appearance reconstruction compared with PSHuman.}

\begin{table}[t]
\scriptsize
	\centering
	\renewcommand\arraystretch{1.25}
	\caption{\revise{Quantitative geometric comparisons among human-centric methods on CityUHuman. The best results are highlighted in \textbf{bold}.}}
	\resizebox{1\linewidth}{!}{
		\begin{tabular}{l|ccc} 
			\toprule
			\multicolumn{1}{c|}{\multirow{2}[0]{*}{\diagbox{Method}{Metric}}} & \multicolumn{3}{c}{CityUHuman}  \\
			& \multicolumn{1}{c}{CD$\downarrow$} & \multicolumn{1}{c}{P2S$\downarrow$} & \multicolumn{1}{c}{Normal$\downarrow$}\\
			\hline
			\hline
			GTA~\cite{zhang2023global}  & 1.151  & 1.071  &2.105     \\ 
			SiTH~\cite{ho2024sith}  & 0.720 & 0.739 & 1.734 \\
            SiFU~\cite{zhang2024sifu}  & 1.273 &1.064 &2.565   \\
            Human-3Diffusion~\cite{xue2024human}  & 0.836 & 0.792 &1.991   \\
            PSHuman~\cite{li2025pshuman}  & 0.754 & 0.788 &1.654  \\
			\hline
			HuGDiffusion & \textbf{0.679}  & \textbf{0.696} & \textbf{1.604}    \\ 
			\bottomrule
	\end{tabular}}
	\label{3dgeometryeval}
\end{table}

\begin{table}[t]
	\scriptsize
	\centering
	\renewcommand\arraystretch{1.25}
	\caption{Quantitative comparisons of GTA, SiTH and HuGDiffusion when providing ground truth and predicted 3D shapes. The best results are highlighted in \textbf{bold}.}
	\resizebox{1.0\linewidth}{!}{
	\begin{tabular}{l|c|c|c|c|c|c} 
		\toprule
		\multicolumn{1}{c|}{\multirow{2}[0]{*}{\diagbox{Method}{Metric}}} & \multicolumn{3}{c|}{Ground Truth Shape}  & \multicolumn{3}{c}{Predicted Shape}    \\
		& \multicolumn{1}{c}{PSNR$\uparrow$} & \multicolumn{1}{c}{SSIM$\uparrow$} & \multicolumn{1}{c|}{LPIPS$\downarrow$} & \multicolumn{1}{c}{PSNR$\uparrow$} & \multicolumn{1}{c}{SSIM$\uparrow$} & \multicolumn{1}{c}{LPIPS$\downarrow$} \\
		\hline
		\hline
		GTA~\cite{zhang2023global}  & 24.82  & 0.930  & 0.059  & 25.78   & 0.919  & 0.085 \\ 
		SiTH~\cite{ho2024sith}  & 26.81 &0.941 &0.048  & 25.36 & 0.919 & 0.083\\
		
		\hline
		HuGDiffusion & \textbf{33.79}  & \textbf{0.971} & \textbf{0.050}  & \textbf{30.03} & \textbf{0.953}   & \textbf{0.065}  \\ 
		\bottomrule
	\end{tabular}}
	\label{3dgtshapetable}
\end{table}
LGM is a generalizable 3DGS model that predicts the 3D Gaussian attributes from the multi-view generated images. However, it occasionally predicts incorrect occluded side images, and the rendered images have low resolution due to consistency issues across different views.  Our HuGDiffusion solves the consistency problem by generating the 3D Gaussian position first. Moreover, owing to our diffusion-based framework, the rendered images are more realistic, compared to LGM. Human-3Diffusion \cite{xue2024human} is \revise{a 3DGS-based methods which specifically trained on human data, however,} Human-3Diffusion cannot preserve the details of the input images, as shown in Fig. \ref{comparesota2}.

SHERF frequently encounters challenges with incorrect poses in SMPL models. While the estimated SMPL models exhibit poses similar to the given view on the 2D image, their poses are frequently incorrect in the 3D space, and SHERF lacks a module to rectify the SMPL models in 3D space. Additionally, SHERF heavily relies on the SMPL model and is unable to render loose clothing, as illustrated in Fig. \ref{inthewildcompare}, where it fails to render the hems of the overalls for the first individual. Furthermore, SHERF directly utilizes the input view to extract image features, which leads to incorrect occluded side information prediction when the occluded side of the human is not symmetrical with the frontal view. Our designed 3D Gaussian position generator and pixel-aligned feature can tackle these issues.

In comparison to SHERF and LGM on in-the-wild images, as shown in Fig. \ref{inthewildcompare}, SHERF fails to recover accurate colors and produces incorrect poses, while also struggling to render loose clothing. LGM, on the other hand, recovers correct colors but cannot preserve face identity or texture details. \revise{In Fig. \ref{inthewild1}, we present several results on in-the-wild images, where our method effectively preserves facial details and reconstructs plausible back-view appearance.}

A geometric comparison of \revise{human-centric methods} in Table \ref{3dgeometryeval} shows that although HuGDiffusion does not prioritize 3D surface recovery, it still delivers the best performance in this regard.

Furthermore, we provide ground truth occupancy or SDF data for GTA and SiTH, and ground truth 3D Gaussian positions for HuGDiffusion to assess the effect of shape correctness, as shown in Table \ref{3dgtshapetable}. Although ground truth data enhances GTA and SiTH, HuGDiffusion achieves even greater gains, showcasing its robustness.

We also refer reviewers to the \textit{Supplementary Material} for the video demo\footnote{\href{https://youtu.be/vadHtUBpEmQ}{https://youtu.be/vadHtUBpEmQ}}.

\subsection{ Ablation Studies}

\begin{figure}[t]
	\begin{center}
		\includegraphics[width=3.4in]{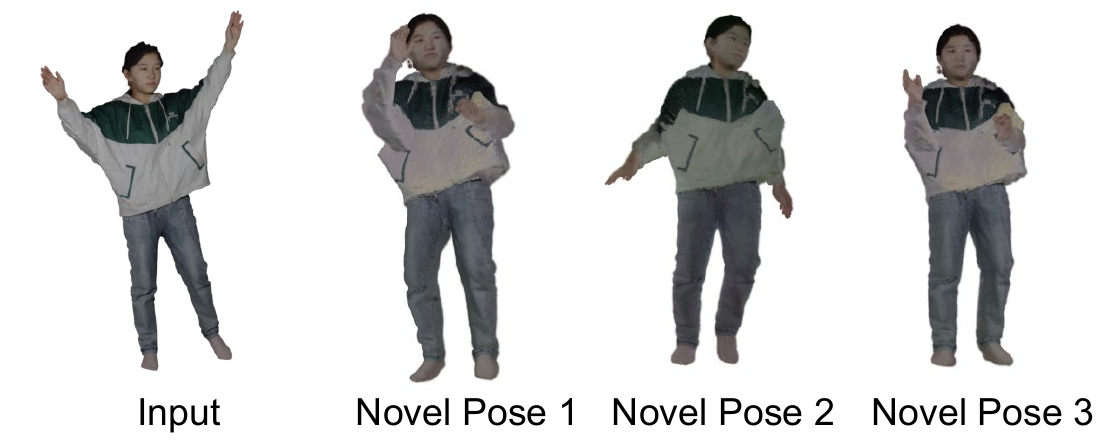}
	\end{center}
	\caption{Novel pose synthesis results of HuGDiffusion.  \textcolor{red}{\faSearch} Zoom in for details.}
	\label{novelpose}
\end{figure}

\textbf{Diffusion-based Model.}
Diffusion models demonstrate superior capabilities in generating unseen appearances compared to regression models. Figure \ref{regressionisbad} provides two examples illustrating this advantage. While regression models can achieve fair numerical performance, as shown in Table \ref{ablation}, they typically fail to accurately reconstruct appearances in occluded areas. For example, in the first example, regardless of whether pixel-supervised or attribute-supervised regression-based models are used, they generate excessive black coloration in the neck and arm areas. In contrast, the diffusion model produces accurate white coloration in these regions. Furthermore, in the second example, both regression-based models generate entirely black heads, which appear incorrect and unnatural, and fail to produce accurate boundaries for the tops and pants. In comparison, the diffusion model generates more realistic and natural appearances in occluded areas. These results highlight the significant advantages of diffusion models in addressing these challenges.

\textbf{Occluded Side Image.}
The predicted occluded side image provides coarse information about unseen areas for HuGDiffusion. When the occluded side image is excluded, as illustrated in Fig. \ref{woback}, HuGDiffusion fails to predict the correct facial identity, resulting in a decline in quantitative performance. These results demonstrate that the occluded side image significantly contributes to enhancing the quality of predictions in unseen areas.

\begin{figure}[t]
	\centering
	\includegraphics[width=3.25in]{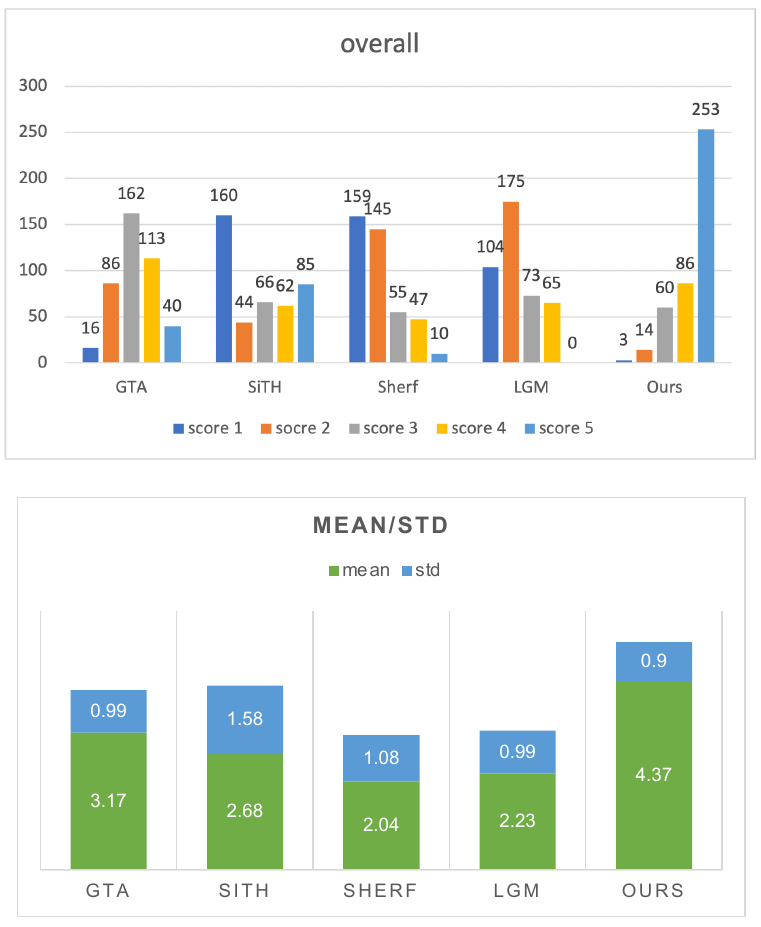}
	\caption{Overall and mean/std results of the subjective evaluation.}
	\label{userstudy2s}
\end{figure}

\textbf{SMPL Semantic Feature.} As presented in Fig. \ref{wosmpl}, when the SMPL semantic feature is not adopted, we observed that the boundary of the neck area became unclear, moreover, there are some red noisy pixels in the image.
 The absence of SMPL semantics constraints causes the network to lack clarity regarding the position of each point on the human body, leading to noisy rendering results and poor numerical performance.

\textbf{Novel Pose Synthesis.} We present several results of novel pose synthesis in Fig. \ref{novelpose}. By blending the 3D Gaussian positions with the SMPL vertices and modifying the SMPL pose, we successfully achieve novel poses. Notably, the results are satisfactory despite the absence of a specifically trained model for this novel pose synthesis task.
\begin{table}[t]\Large
	\centering
	\renewcommand\arraystretch{1.25}
	\caption{The results of ablation studies on Thuman.  R: regression-based model. D: diffusion-based model.}
	\label{ablation}
	\scriptsize
	\resizebox{1.0\linewidth}{!}{\begin{tabular}{ccc|cc|ccc}
			\toprule
			\multicolumn{3}{c|}{}                                                                          & \begin{tabular}[c]{@{}c@{}}Back\\ Image\end{tabular} & \begin{tabular}[c]{@{}c@{}}Smpl\\ Semantic\end{tabular} & PSNR$\uparrow$ & SSIM$\uparrow$ & LPIPS$\downarrow$ \\ \hline  
			\multicolumn{3}{c|}{Pixle-Level}                                                                  &\faCheck&\faCheck                                                       &29.02  &0.953 &0.070     \\ \hline
			\multicolumn{1}{c|}{\multirow{4}{*}{ \makecell{Attribute- \\ Level} }} & \multicolumn{2}{c|}{R}                 &\faCheck&\faCheck                                                        & 29.71 & 0.951 & 0.065  \\ \cline{2-8} 
			\multicolumn{1}{c|}{}                        & \multicolumn{2}{c|}{\multirow{3}{*}{D}} & \faTimes                                                    & \faTimes                                                       & 28.44   & 0.948   & 0.074    \\ 
			\multicolumn{1}{c|}{}                        & \multicolumn{2}{c|}{}                           &\faCheck&\faTimes                                                    & 29.63   & 0.950    & 0.070     \\  
			\multicolumn{1}{c|}{}                        & \multicolumn{2}{c|}{}                           &\faCheck&\faCheck                                                      & 30.03    & 0.953    & 0.065     \\ \bottomrule
	\end{tabular}}
\end{table}

\subsection{Perceptual  Evaluation}

We conducted a Perceptual evaluation to compare various methods quantitatively. Specifically, we engaged 52 participants, including undergraduate students, postgraduate students from diverse research backgrounds, and industry professionals, to assess 8 different human bodies. For each human body, we presented three images generated by different methods and asked the participants to provide scores ranging from 1 to 5, reflecting the quality of the generated shapes. The rating scale was as follows: 1 - poor, 2 - below average, 3 - average, 4 - good, and 5 - excellent.

Fig. \ref{userstudy2s} shows the results of the subjective evaluation, including the overall scores, mean values, and standard deviations (std) of the scores. It can be observed that our HuGDiffusion achieves the highest mean score and the lowest std value.

\section{Conclusion}
\label{sectionconclusion}

We introduced HuGDiffusion, an innovative 3D Gaussian attribute diffusion framework for novel view synthesis from single-view human images. The approach employs a two-stage workflow to construct precise 3D Gaussian attributes, enabling a diffusion training process driven by attribute-level supervision. Utilizing a PointNet++-based architecture, the framework effectively denoises the 3D Gaussian attributes to generate accurate and plausible appearances for occluded areas. Additionally, human-centric features were integrated as conditions to enhance the training process. Extensive experimental evaluations reveal that HuGDiffusion consistently outperforms state-of-the-art methods across both quantitative metrics and qualitative assessments. 

\revisesec{\noindent\textbf{\textit{Limitations and Future Works.}} Currently, our HuGDiffusion still produces blurriness in certain unseen regions due to the insufficient number of points in the generated complete 3D human point cloud. To further improve the appearance quality in unseen regions, we plan to investigate more powerful point set architectures for generating much denser and higher-quality human point clouds. Besides, it is also promising to integrate high-fidelity image generative models to supplement more reliable visual cues for unseen regions.}

\bibliographystyle{ieee_fullname}
\bibliography{egbib}

\begin{thebibliography}{10}\itemsep=-1pt

\bibitem{brooks2023instructpix2pix}
Tim Brooks, Aleksander Holynski, and Alexei~A Efros.
\newblock Instructpix2pix: Learning to follow image editing instructions.
\newblock In {\em Proc. CVPR}, pages 18392--18402, 2023.

\bibitem{cheng2025pvnet}
Xianjing Cheng, Lintai Wu, Zuowen Wang, Junhui Hou, Jie Wen, and Yong Xu.
\newblock Pvnet: Point-voxel interaction lidar scene upsampling via diffusion
  models.
\newblock {\em arXiv preprint arXiv:2508.17050}, 2025.

\bibitem{gao2022mps}
Xiangjun Gao, Jiaolong Yang, Jongyoo Kim, Sida Peng, Zicheng Liu, and Xin Tong.
\newblock Mps-nerf: Generalizable 3d human rendering from multiview images.
\newblock {\em IEEE TPAMI}, 2022.

\bibitem{han2023high}
Sang-Hun Han, Min-Gyu Park, Ju~Hong Yoon, Ju-Mi Kang, Young-Jae Park, and
  Hae-Gon Jeon.
\newblock High-fidelity 3d human digitization from single 2k resolution images.
\newblock In {\em Proc. CVPR}, pages 12869--12879, 2023.

\bibitem{he2024magicman}
Xu He, Xiaoyu Li, Di Kang, Jiangnan Ye, Chaopeng Zhang, Liyang Chen, Xiangjun
  Gao, Han Zhang, Zhiyong Wu, and Haolin Zhuang.
\newblock Magicman: Generative novel view synthesis of humans with 3d-aware
  diffusion and iterative refinement.
\newblock {\em arXiv preprint arXiv:2408.14211}, 2024.

\bibitem{ho2023learning}
Hsuan-I Ho, Lixin Xue, Jie Song, and Otmar Hilliges.
\newblock Learning locally editable virtual humans.
\newblock In {\em Proc. CVPR}, pages 21024--21035, 2023.

\bibitem{ho2024sith}
I Ho, Jie Song, Otmar Hilliges, et~al.
\newblock Sith: Single-view textured human reconstruction with
  image-conditioned diffusion.
\newblock In {\em Proc. CVPR}, pages 538--549, 2024.

\bibitem{hou2024global}
Jinhui Hou, Zhiyu Zhu, Junhui Hou, Hui Liu, Huanqiang Zeng, and Hui Yuan.
\newblock Global structure-aware diffusion process for low-light image
  enhancement.
\newblock In {\em Proc. NeurIPS}, 2024.

\bibitem{hu2023sherf}
Shoukang Hu, Fangzhou Hong, Liang Pan, Haiyi Mei, Lei Yang, and Ziwei Liu.
\newblock {SHERF}: Generalizable human nerf from a single image.
\newblock In {\em Proc. ICCV}, pages 9352--9364, 2023.

\bibitem{hu2024gauhuman}
Shoukang Hu, Tao Hu, and Ziwei Liu.
\newblock Gauhuman: Articulated gaussian splatting from monocular human videos.
\newblock In {\em Proc. CVPR}, pages 20418--20431, 2024.

\bibitem{jiang2022neuman}
Wei Jiang, Kwang~Moo Yi, Golnoosh Samei, Oncel Tuzel, and Anurag Ranjan.
\newblock Neuman: Neural human radiance field from a single video.
\newblock In {\em Proc. ECCV}, pages 402--418, 2022.

\bibitem{kerbl20233d}
Bernhard Kerbl, Georgios Kopanas, Thomas Leimk{\"u}hler, and George Drettakis.
\newblock 3d gaussian splatting for real-time radiance field rendering.
\newblock {\em ACM TOG}, 42(4):139--1, 2023.

\bibitem{kocabas2024hugs}
Muhammed Kocabas, Jen-Hao~Rick Chang, James Gabriel, Oncel Tuzel, and Anurag
  Ranjan.
\newblock Hugs: Human gaussian splats.
\newblock In {\em Proc. CVPR}, pages 505--515, 2024.

\bibitem{li2025pshuman}
Peng Li, Wangguandong Zheng, Yuan Liu, Tao Yu, Yangguang Li, Xingqun Qi,
  Xiaowei Chi, Siyu Xia, Yan-Pei Cao, Wei Xue, et~al.
\newblock Pshuman: Photorealistic single-image 3d human reconstruction using
  cross-scale multiview diffusion and explicit remeshing.
\newblock In {\em Proc. CVPR}, pages 16008--16018, 2025.

\bibitem{liu2023zero}
Ruoshi Liu, Rundi Wu, Basile Van~Hoorick, Pavel Tokmakov, Sergey Zakharov, and
  Carl Vondrick.
\newblock Zero-1-to-3: Zero-shot one image to 3d object.
\newblock In {\em Proc. ICCV}, pages 9298--9309, 2023.

\bibitem{liu2024fast}
Tianqi Liu, Guangcong Wang, Shoukang Hu, Liao Shen, Xinyi Ye, Yuhang Zang,
  Zhiguo Cao, Wei Li, and Ziwei Liu.
\newblock Fast generalizable gaussian splatting reconstruction from multi-view
  stereo.
\newblock In {\em Proc. ECCV}, 2024.

\bibitem{luo2021diffusion}
Shitong Luo and Wei Hu.
\newblock Diffusion probabilistic models for 3d point cloud generation.
\newblock In {\em Proc. CVPR}, pages 2837--2845, 2021.

\bibitem{lyu2022a}
Zhaoyang Lyu, Zhifeng Kong, Xudong XU, Liang Pan, and Dahua Lin.
\newblock A conditional point diffusion-refinement paradigm for 3d point cloud
  completion.
\newblock In {\em Proc. ICLR}, 2022.

\bibitem{lyu2023controllable}
Zhaoyang Lyu, Jinyi Wang, Yuwei An, Ya Zhang, Dahua Lin, and Bo Dai.
\newblock Controllable mesh generation through sparse latent point diffusion
  models.
\newblock In {\em Proc. CVPR}, pages 271--280, 2023.

\bibitem{melas2023pc2}
Luke Melas-Kyriazi, Christian Rupprecht, and Andrea Vedaldi.
\newblock Pc2: Projection-conditioned point cloud diffusion for single-image 3d
  reconstruction.
\newblock In {\em Proc. CVPR}, pages 12923--12932, 2023.

\bibitem{mildenhall2020nerf}
Ben Mildenhall, Pratul~P Srinivasan, Matthew Tancik, Jonathan~T Barron, Ravi
  Ramamoorthi, and Ren Ng.
\newblock {NeRF}: Representing scenes as neural radiance fields for view
  synthesis.
\newblock In {\em Proc. ECCV}, pages 99--106, 2020.

\bibitem{mu2023actorsnerf}
Jiteng Mu, Shen Sang, Nuno Vasconcelos, and Xiaolong Wang.
\newblock Actorsnerf: Animatable few-shot human rendering with generalizable
  nerfs.
\newblock In {\em Proc. ICCV}, pages 18391--18401, 2023.

\bibitem{palfinger2022continuous}
Werner Palfinger.
\newblock Continuous remeshing for inverse rendering.
\newblock {\em Computer Animation and Virtual Worlds}, 33(5):e2101, 2022.

\bibitem{park2019deepsdf}
Jeong~Joon Park, Peter Florence, Julian Straub, Richard Newcombe, and Steven
  Lovegrove.
\newblock {DeepSDF}: Learning continuous signed distance functions for shape
  representation.
\newblock In {\em Proc. CVPR}, pages 165--174, 2019.

\bibitem{patni2024ecodepth}
Suraj Patni, Aradhye Agarwal, and Chetan Arora.
\newblock {ECoDepth}: Effective conditioning of diffusion models for monocular
  depth estimation.
\newblock In {\em Proc. CVPR}, pages 28285--28295, 2024.

\bibitem{peng2021neural}
Sida Peng, Yuanqing Zhang, Yinghao Xu, Qianqian Wang, Qing Shuai, Hujun Bao,
  and Xiaowei Zhou.
\newblock Neural body: Implicit neural representations with structured latent
  codes for novel view synthesis of dynamic humans.
\newblock In {\em Proc. CVPR}, pages 9054--9063, 2021.

\bibitem{rahaman2019spectral}
Nasim Rahaman, Aristide Baratin, Devansh Arpit, Felix Draxler, Min Lin, Fred
  Hamprecht, Yoshua Bengio, and Aaron Courville.
\newblock On the spectral bias of neural networks.
\newblock In {\em Proc. ICML}, pages 5301--5310, 2019.

\bibitem{rai2025uvgs}
Aashish Rai, Dilin Wang, Mihir Jain, Nikolaos Sarafianos, Kefan Chen, Srinath
  Sridhar, and Aayush Prakash.
\newblock Uvgs: Reimagining unstructured 3d gaussian splatting using uv
  mapping.
\newblock In {\em Proceedings of the Computer Vision and Pattern Recognition
  Conference}, pages 5927--5937, 2025.

\bibitem{ren2025ddm}
Siyu Ren, Junhui Hou, Xiaodong Chen, Hongkai Xiong, and Wenping Wang.
\newblock Ddm: A metric for comparing 3d shapes using directional distance
  fields.
\newblock {\em IEEE TPAMI}, 2025.

\bibitem{roessle2024l3dg}
Barbara Roessle, Norman M{\"u}ller, Lorenzo Porzi, Samuel Rota~Bul{\`o}, Peter
  Kontschieder, Angela Dai, and Matthias Nie{\ss}ner.
\newblock L3dg: Latent 3d gaussian diffusion.
\newblock In {\em Proc. SIGGRAPH Asia}, pages 1--11, 2024.

\bibitem{rombach2022high}
Robin Rombach, Andreas Blattmann, Dominik Lorenz, Patrick Esser, and Bj{\"o}rn
  Ommer.
\newblock High-resolution image synthesis with latent diffusion models.
\newblock In {\em Proc. CVPR}, pages 10684--10695, 2022.

\bibitem{saharia2022image}
Chitwan Saharia, Jonathan Ho, William Chan, Tim Salimans, David~J Fleet, and
  Mohammad Norouzi.
\newblock Image super-resolution via iterative refinement.
\newblock {\em IEEE TPAMI}, 45(4):4713--4726, 2022.

\bibitem{saito2019pifu}
Shunsuke Saito, Zeng Huang, Ryota Natsume, Shigeo Morishima, Angjoo Kanazawa,
  and Hao Li.
\newblock Pifu: Pixel-aligned implicit function for high-resolution clothed
  human digitization.
\newblock In {\em Proc. ICCV}, pages 2304--2314, 2019.

\bibitem{saito2020pifuhd}
Shunsuke Saito, Tomas Simon, Jason Saragih, and Hanbyul Joo.
\newblock Pifuhd: Multi-level pixel-aligned implicit function for
  high-resolution 3d human digitization.
\newblock In {\em Proc. CVPR}, pages 84--93, 2020.

\bibitem{tang2024lgm}
Jiaxiang Tang, Zhaoxi Chen, Xiaokang Chen, Tengfei Wang, Gang Zeng, and Ziwei
  Liu.
\newblock Lgm: Large multi-view gaussian model for high-resolution 3d content
  creation.
\newblock In {\em Proc. ECCV}, pages 1--18, 2024.

\bibitem{tang2025human}
Yingzhi Tang, Qijian Zhang, Yebin Liu, and Junhui Hou.
\newblock Human as points: Explicit point-based 3d human reconstruction from
  single-view rgb images.
\newblock {\em IEEE TPAMI}, 2025.

\bibitem{tian2023recovering}
Yating Tian, Hongwen Zhang, Yebin Liu, and Limin Wang.
\newblock Recovering 3d human mesh from monocular images: A survey.
\newblock {\em IEEE TPAMI}, 2023.

\bibitem{vahdat2022lion}
Arash Vahdat, Francis Williams, Zan Gojcic, Or Litany, Sanja Fidler, Karsten
  Kreis, et~al.
\newblock {LION}: Latent point diffusion models for 3d shape generation.
\newblock In {\em Proc. NeurIPS}, pages 10021--10039, 2022.

\bibitem{wang2024freesplat}
Yunsong Wang, Tianxin Huang, Hanlin Chen, and Gim~Hee Lee.
\newblock Freesplat: Generalizable 3d gaussian splatting towards free-view
  synthesis of indoor scenes.
\newblock In {\em Proc. NeurIPS}, 2024.

\bibitem{weng2022humannerf}
Chung-Yi Weng, Brian Curless, Pratul~P Srinivasan, Jonathan~T Barron, and Ira
  Kemelmacher-Shlizerman.
\newblock Humannerf: Free-viewpoint rendering of moving people from monocular
  video.
\newblock In {\em Proc. CVPR}, pages 16210--16220, 2022.

\bibitem{wu2025unsupervised}
Lintai Wu, Xianjing Cheng, Yong Xu, Huanqiang Zeng, and Junhui Hou.
\newblock Unsupervised 3d point cloud completion via multi-view adversarial
  learning.
\newblock {\em IEEE TVCG}, 2025.

\bibitem{wu20243d}
Lintai Wu, Junhui Hou, Linqi Song, and Yong Xu.
\newblock 3d shape completion on unseen categories: A weakly-supervised
  approach.
\newblock {\em IEEE TVCG}, 2024.

\bibitem{xiu2022icon}
Yuliang Xiu, Jinlong Yang, Dimitrios Tzionas, and Michael~J Black.
\newblock {ICON}: Implicit clothed humans obtained from normals.
\newblock In {\em Proc. CVPR}, pages 13286--13296, 2022.

\bibitem{xu2021h}
Hongyi Xu, Thiemo Alldieck, and Cristian Sminchisescu.
\newblock H-nerf: Neural radiance fields for rendering and temporal
  reconstruction of humans in motion.
\newblock In {\em Proc. NeurIPS}, pages 14955--14966, 2021.

\bibitem{xue2024human}
Yuxuan Xue, Xianghui Xie, Riccardo Marin, and Gerard Pons-Moll.
\newblock Human 3diffusion: Realistic avatar creation via explicit 3d
  consistent diffusion models.
\newblock In {\em Proc. NeurIPS}, 2024.

\bibitem{xue2025gen}
Yuxuan Xue, Xianghui Xie, Riccardo Marin, and Gerard Pons-Moll.
\newblock Gen-3diffusion: Realistic image-to-3d generation via 2d \& 3d
  diffusion synergy.
\newblock {\em IEEE TPAMI}, 2025.

\bibitem{yang2024magic}
Fan Yang, Jianfeng Zhang, Yichun Shi, Bowen Chen, Chenxu Zhang, Huichao Zhang,
  Xiaofeng Yang, Jiashi Feng, and Guosheng Lin.
\newblock Magic-boost: Boost 3d generation with mutli-view conditioned
  diffusion.
\newblock {\em arXiv preprint arXiv:2404.06429}, 2024.

\bibitem{yu2021pixelnerf}
Alex Yu, Vickie Ye, Matthew Tancik, and Angjoo Kanazawa.
\newblock pixelnerf: Neural radiance fields from one or few images.
\newblock In {\em Proc. CVPR}, pages 4578--4587, 2021.

\bibitem{yu2021function4d}
Tao Yu, Zerong Zheng, Kaiwen Guo, Pengpeng Liu, Qionghai Dai, and Yebin Liu.
\newblock Function4d: Real-time human volumetric capture from very sparse
  consumer rgbd sensors.
\newblock In {\em Proc. CVPR}, pages 5746--5756, 2021.

\bibitem{zeng2024dynamic}
Yiming Zeng, Junhui Hou, Qijian Zhang, Siyu Ren, and Wenping Wang.
\newblock Dynamic 3d point cloud sequences as 2d videos.
\newblock {\em IEEE TPAMI}, 46(12):9371--9386, 2024.

\bibitem{zhanggaussiancube}
Bowen Zhang, Yiji Cheng, Jiaolong Yang, Chunyu Wang, Feng Zhao, Yansong Tang,
  Dong Chen, and Baining Guo.
\newblock Gaussiancube: A structured and explicit radiance representation for
  3d generative modeling.
\newblock In {\em Proc. NeurIPS}, 2024.

\bibitem{zhang2023adding}
Lvmin Zhang, Anyi Rao, and Maneesh Agrawala.
\newblock Adding conditional control to text-to-image diffusion models.
\newblock In {\em Proc. ICCV}, pages 3836--3847, 2023.

\bibitem{zhang2023flattening}
Qijian Zhang, Junhui Hou, Yue Qian, Yiming Zeng, Juyong Zhang, and Ying He.
\newblock Flattening-net: Deep regular 2d representation for 3d point cloud
  analysis.
\newblock {\em IEEE TPAMI}, 45(8):9726--9742, 2023.

\bibitem{zhang2023global}
Zechuan Zhang, Li Sun, Zongxin Yang, Ling Chen, and Yi Yang.
\newblock Global-correlated 3d-decoupling transformer for clothed avatar
  reconstruction.
\newblock In {\em Proc. NeurIPS}, 2023.

\bibitem{zhang2024sifu}
Zechuan Zhang, Zongxin Yang, and Yi Yang.
\newblock Sifu: Side-view conditioned implicit function for real-world usable
  clothed human reconstruction.
\newblock In {\em Proc. CVPR}, pages 9936--9947, 2024.

\bibitem{zhao2022humannerf}
Fuqiang Zhao, Wei Yang, Jiakai Zhang, Pei Lin, Yingliang Zhang, Jingyi Yu, and
  Lan Xu.
\newblock Humannerf: Efficiently generated human radiance field from sparse
  inputs.
\newblock In {\em Proc. CVPR}, pages 7743--7753, 2022.

\bibitem{zheng2024gps}
Shunyuan Zheng, Boyao Zhou, Ruizhi Shao, Boning Liu, Shengping Zhang, Liqiang
  Nie, and Yebin Liu.
\newblock Gps-gaussian: Generalizable pixel-wise 3d gaussian splatting for
  real-time human novel view synthesis.
\newblock In {\em Proc. CVPR}, pages 19680--19690, 2024.

\bibitem{zheng2021pamir}
Zerong Zheng, Tao Yu, Yebin Liu, and Qionghai Dai.
\newblock Pamir: Parametric model-conditioned implicit representation for
  image-based human reconstruction.
\newblock {\em IEEE TPAMI}, 44(6):3170--3184, 2021.

\bibitem{zhou2024diffgs}
Junsheng Zhou, Weiqi Zhang, and Yu-Shen Liu.
\newblock Diffgs: Functional gaussian splatting diffusion.
\newblock In {\em Proc. NeurIPS}, 2024.

\bibitem{zhou20213d}
Linqi Zhou, Yilun Du, and Jiajun Wu.
\newblock 3d shape generation and completion through point-voxel diffusion.
\newblock In {\em Proc. ICCV}, pages 5826--5835, 2021.

\bibitem{zou2024triplane}
Zi-Xin Zou, Zhipeng Yu, Yuan-Chen Guo, Yangguang Li, Ding Liang, Yan-Pei Cao,
  and Song-Hai Zhang.
\newblock Triplane meets gaussian splatting: Fast and generalizable single-view
  3d reconstruction with transformers.
\newblock In {\em Proc. CVPR}, pages 10324--10335, 2024.

\end{thebibliography}

\newpage
\renewcommand{\thesection}{S\arabic{section}}  
\renewcommand{\thesubsection}{S\arabic{section}.\arabic{subsection}} 

\setcounter{section}{0} 

\addcontentsline{toc}{section}{Supplementary Materials} 
\renewcommand{\thefigure}{S\arabic{figure}}
\renewcommand{\thetable}{S\arabic{table}}
\setcounter{figure}{0}
\setcounter{table}{0}

\onecolumn
\title{\vspace{-2em}Supplementary Material for \\[0.3em]
\textbf{HuGDiffusion: Generalizable Single-Image Human Rendering via 3D Gaussian Diffusion}}
\date{}
\author{}

\maketitle

\section{Experiments on Larger Training Dataset}
We trained HuGDiffusion with an expanded dataset by supplementing 1,600 scans from the THuman2.1 dataset, leading to a total of 2,080 samples. The results in Table~\ref{r2q1t} demonstrate consistent performance improvements in all metrics.

\begin{table*}[h]
    \scriptsize
    \centering
    \renewcommand\arraystretch{1.25}
    \caption{Quantitative comparisons on Thuman, CityuHuman, 2K2K, and CustomHuman datasets. The best results are highlighted in \textbf{bold}. $\uparrow$: the higher the better. $\downarrow$: the lower the better.}
    \resizebox{0.97\linewidth}{!}{\begin{tabular}{l|c|c|c|c|c|c|c|c|c|c|c|c} 
            \toprule
            \multicolumn{1}{c|}{\multirow{2}[0]{*}{\diagbox{Method}{Metric}}} & \multicolumn{3}{c|}{Thuman} & \multicolumn{3}{c|}{CityuHuman} & \multicolumn{3}{c|}{2K2K} & \multicolumn{3}{c}{CustomHuman}    \\
            & \multicolumn{1}{c}{PSNR$\uparrow$} & \multicolumn{1}{c}{SSIM$\uparrow$} & \multicolumn{1}{c|}{LPIPS$\downarrow$} & \multicolumn{1}{c}{PSNR$\uparrow$} & \multicolumn{1}{c}{SSIM$\uparrow$} & \multicolumn{1}{c|}{LPIPS$\downarrow$} & \multicolumn{1}{c}{PSNR$\uparrow$} & \multicolumn{1}{c}{SSIM$\uparrow$} & \multicolumn{1}{c|}{LPIPS$\downarrow$}   & \multicolumn{1}{c}{PSNR$\uparrow$} & \multicolumn{1}{c}{SSIM$\uparrow$} & \multicolumn{1}{c}{LPIPS$\downarrow$}\\
            \hline
            \hline
            GTA ~\cite{zhang2023global} & 25.78   & 0.919  & 0.085  & 27.41  & 0.923  & 0.075& 24.15  & 0.921  & 0.080 &28.86 &0.920&0.088 \\ 
            SiTH ~\cite{ho2024sith} & 25.36 & 0.919 & 0.083 & 29.21 &0.934 &0.067 & 24.30
            & 0.920 & 0.076 &26.47 &0.911&0.095 \\
            LGM ~\cite{tang2024lgm} & 25.13  & 0.915 & 0.096 & 29.78   & 0.941 & 0.074 &27.99 & 0.938&0.071 &31.91& 0.944 &0.077 \\ 
            SHERF ~\cite{hu2023sherf} & 26.57   & 0.927   & 0.081 & 30.13   & 0.942   & 0.067 &27.29&0.931&0.072&27.88&0.916&0.096\\ 
            SIFU \cite{zhang2024sifu} & 23.16
            & 0.904 & 0.102 & 26.46
            & 0.917 & 0.087& 24.30
            & 0.920 & 0.076& 29.62 & 0.928 & 0.092\\
            Human-3Diffusion~\cite{xue2024human} & 27.06   & 0.934   & 0.079 & 30.48   & 0.944   & 0.068 &29.05&0.942&0.062&33.75&0.952&0.067\\
            PSHuman~\cite{li2025pshuman} & 25.34  & 0.910   & 0.084 & 27.82   & 0.925   & 0.071 & 24.72 & 0.917 & 0.067 & 30.26 & 0.931 & 0.082\\
            \hline
            HuGDiffusion \textcolor{gray}{Neural} & 29.70  & 0.950   & 0.069 & 32.39   & 0.953   & 0.064 & 30.18 & 0.947 & 0.062 & 34.64 & 0.953 & 0.059 \\
            HuGDiffusion \textcolor{gray}{Joint} & \textbf{30.03} & \textbf{0.953}   & \textbf{0.065} & \textbf{32.47}  & \textbf{0.954} & \textbf{0.062}  & \textbf{30.64}  & \textbf{0.949} & \textbf{0.060} & \textbf{34.82}  & \textbf{0.958} & \textbf{0.055} \\ 
            HuGDiffusion \textcolor{gray}{More Data} & \textcolor{gray}{\textbf{30.21}} & \textcolor{gray}{\textbf{0.955}}   & \textcolor{gray}{\textbf{0.065}} & \textcolor{gray}{\textbf{32.69}}  & \textcolor{gray}{\textbf{0.956}} & \textcolor{gray}{\textbf{0.061}}  & \textcolor{gray}{\textbf{30.89}}  & \textcolor{gray}{\textbf{0.950}} & \textcolor{gray}{\textbf{0.059}} & \textcolor{gray}{\textbf{35.20}}  & \textcolor{gray}{\textbf{0.961}} & \textcolor{gray}{\textbf{0.054}} \\ 
            \bottomrule
    \end{tabular}}
    \label{r2q1t}
\end{table*}

\section{Alternative Attribute Regularization Strategies}
\subsection{Gaussian Scale Clipping} 
\begin{figure*}[t!]
    \centering
    \includegraphics[width=6.8in]{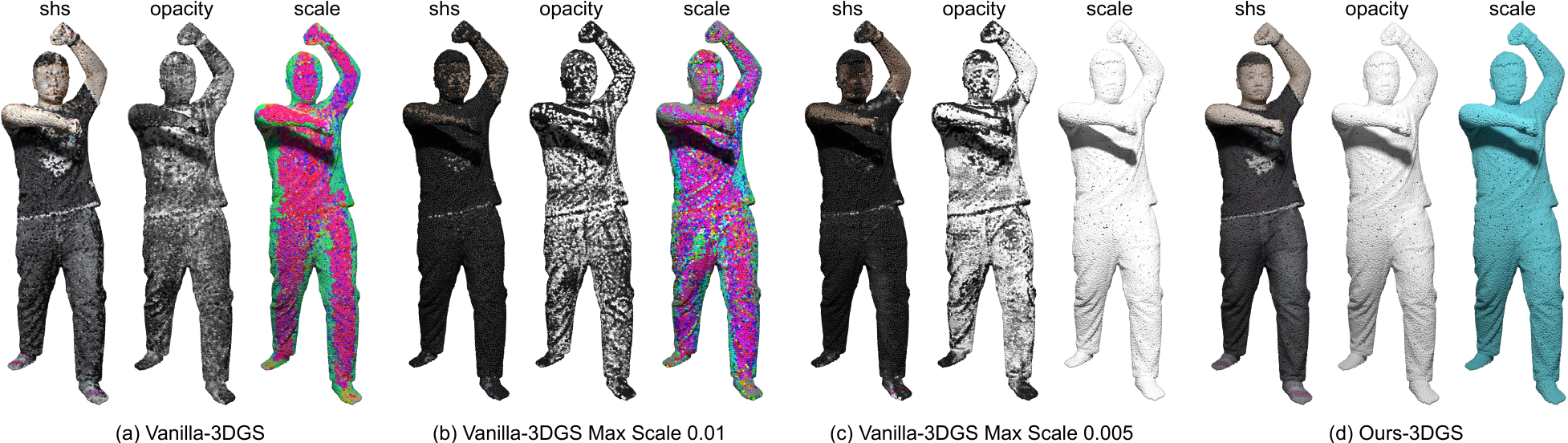}
    \caption{  Visualization of spherical harmonic, opacity, and scale for vanilla-3DGS and our proxy ground-truth 3D Gaussian attributes.  \textcolor{red}{\faSearch} Zoom in for details.}
    \label{r3scaleclamp}
\end{figure*}

We accordingly conducted experiments by adopting the same setting as DiffGS \cite{zhou2024diffgs}, where scales are clipped to a maximum of 0.01 to avoid abnormal Gaussians. However, this simple regularization only prevents the emergence of extremely large Gaussians in 3DGS, without reducing the randomness inherent in the 3DGS attributes. As shown in Fig. \ref{r3scaleclamp} (b), the randomness remains evident in the 3DGS attributes, similar to what can be observed in Fig. \ref{r3scaleclamp} (a). As shown in Fig. \ref{r3scaleclamp} (c), we also tested clamping the  maximum scale to 0.005; while this results in more regular scales, such regularization still fails to adequately overcome randomness. Consequently, such 3DGS attributes are unsuitable as ground truth for training HuGDiffusion, as their randomness prevents the loss from converging and \textbf{ultimately leads to training failure}.
\subsection{MLP for Individual Fitting} 
\begin{figure*}[t!]
    \centering
    \includegraphics[width=3.5in]{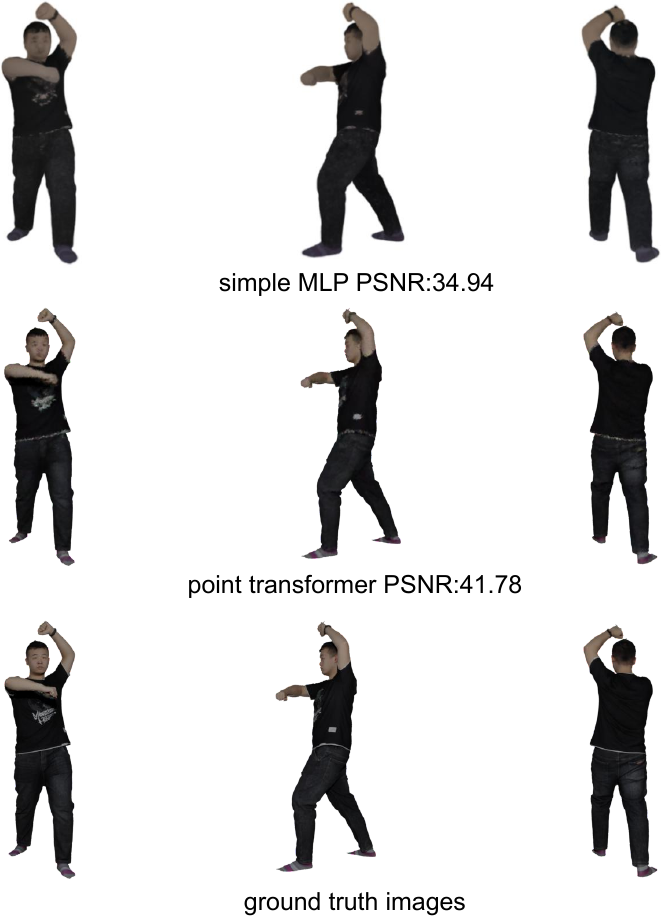}
    \caption{  The constructed results on 2K2K and CustomHuman
    datasets.  \textcolor{red}{\faSearch} Zoom in for details.}
    \label{simplemlp}
\end{figure*}

We also provide individual fitting results with simple MLP instead of point transformer. Simple MLPs lack the awareness of geometric information and fail to capture the local features of point clouds. As a result, it typically learns low-quality 3DGS attributes, which in turn leads to blurred rendering results, as illustrated in Fig. \ref{simplemlp}. Therefore, we adopted the more powerful point cloud transformer architecture. 

\section{Alternative Appearance Representations}
In HuGDiffusion, although human appearance is mostly dominated by diffuse colors, in practice the more expressive representation format of spherical harmonics (SHs) typically shows better performance. Below we particularly evaluate this issue by conducting targeted experiments.

\begin{figure*}[t!]
    \centering
    \includegraphics[width=6in]{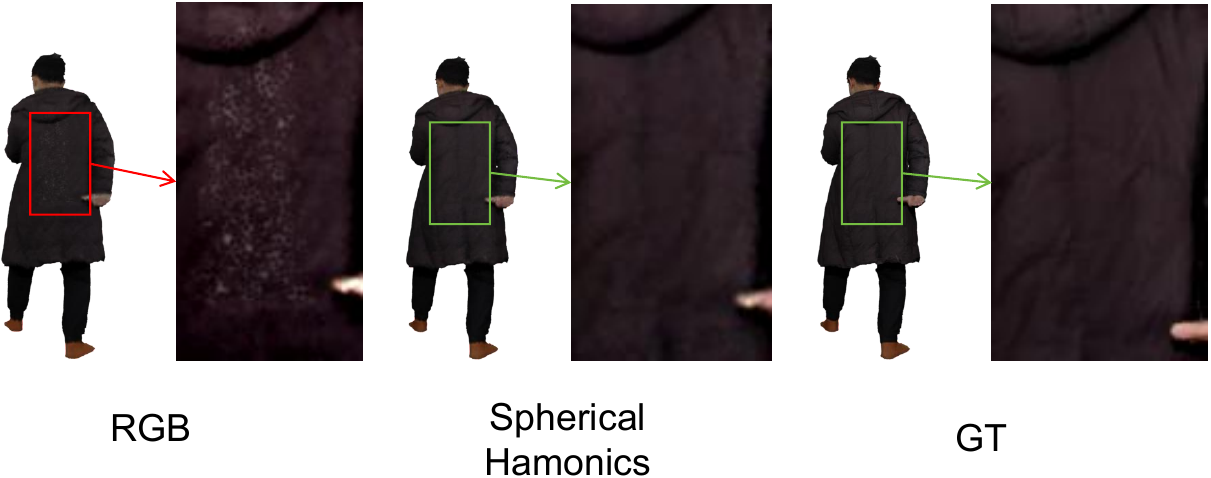}
    \caption{The overfitting results of RGB and Spherical Harmonics.  \textcolor{red}{\faSearch} Zoom in for details.}
    \label{r3q3fig}
\end{figure*}

\noindent First, we evaluate the expressiveness of the two formats through an overfitting experiment on ten subjects in the THuman2 dataset (with IDs from 0000 to 0009). As shown in Table \ref{r3q3table}, replacing SHs with RGB leads to proniment PSNR decrease. As visually compared in Figure \ref{r3q3fig}, using RGB for appearance modeling can cause more image noises. These quantitative and qualitative results demonstrate that spherical harmonics are more suitable for accurate appearance modeling for presenting subtle shading effects and fine-grained appearance variations. Furthermore, we quantitatively compare the impacts of different appearance modeling methods on our final performance. As shown in Table \ref{r3q4table}, when switching to RGB for our ground-truth construction, we observe consistent performance drops, demonstrating that spherical harmonics are also more effective than the simple RGB representation format in the actual training process. \\

\begin{table*}[t!]
    \scriptsize
    \centering
    \renewcommand\arraystretch{1.25}
    \caption{The PSNR of different person.}
    \resizebox{0.97\linewidth}{!}{\begin{tabular}{l|cccccccccc} 
            \toprule
            \multicolumn{1}{c|}{{\diagbox{Method}{Person}}} & \multicolumn{1}{c}{0000} & \multicolumn{1}{c}{0001} & \multicolumn{1}{c}{0002} & \multicolumn{1}{c}{0003}   & \multicolumn{1}{c}{0004}& \multicolumn{1}{c}{0005}& \multicolumn{1}{c}{0006}& \multicolumn{1}{c}{0007}& \multicolumn{1}{c}{0008}& \multicolumn{1}{c}{0009}\\
            
            \hline
            \hline
            RGB  & 38.34   & 37.92  & 37.04 & 36.62  & 37.05  & 37.78& 36.35  & 34.50  & 41.70 & 40.13  \\ 
            Spherical Harmonics & 41.78 & 42.53 & 42.31 & 41.12 &42.94 &41.96 & 41.79
            & 39.01 & 46.35 & 44.07  \\
            
            \bottomrule
    \end{tabular}}
    \label{r3q3table}
\end{table*}

\begin{table*}[t!]
    \scriptsize
    \centering
    \renewcommand\arraystretch{1.25}
    \caption{The quantitative results of different ground truth 3DGS attributes on THuman.}
    \resizebox{0.4\linewidth}{!}{\begin{tabular}{l|cccccccccc} 
            \toprule
            \multicolumn{1}{c|}{{\diagbox{GT}{Metric}}} & \multicolumn{1}{c}{PSNR } & \multicolumn{1}{c}{SSIM }& \multicolumn{1}{c}{LPIPS } \\
            
            \hline
            \hline
            RGB  & 29.44  & 0.949  &0.069  \\ 
            Spherical Harmonics & 30.03 & 0.953 & 0.065   \\
            
            \bottomrule
    \end{tabular}}
    \label{r3q4table}
\end{table*}

\section{Network Structures}

\begin{figure*}[t!]
    \centering
    \setlength{\abovecaptionskip}{0.1cm}
    \includegraphics[width=5.4in]{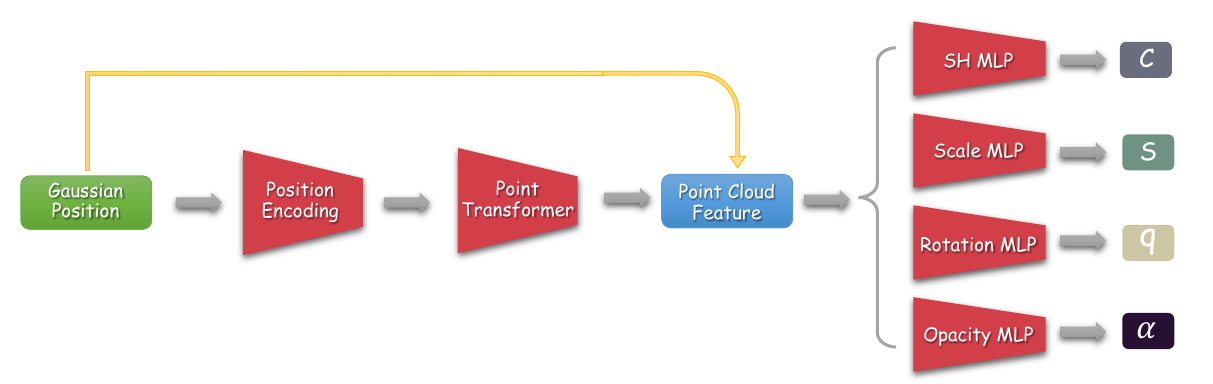}
    \caption{The architecture of point transformer \( \Theta_1 \).}
    \vspace{-3mm}
    \label{pointransformer1}
\end{figure*}

\begin{figure*}[t!]
    \centering
    \setlength{\abovecaptionskip}{0.1cm}
    \includegraphics[width=5.4in]{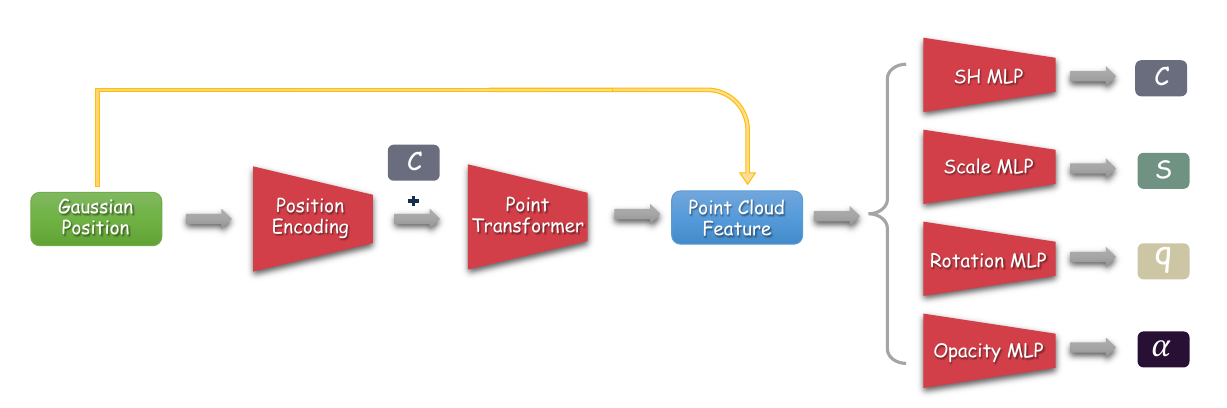}
    \caption{The architecture of point transformer \( \Theta_2 \).}
    \vspace{-3mm}
    \label{pointransformer2}
\end{figure*}

\begin{figure*}[t!]
    \centering
    \setlength{\abovecaptionskip}{0.1cm}
    \includegraphics[width=5.4in]{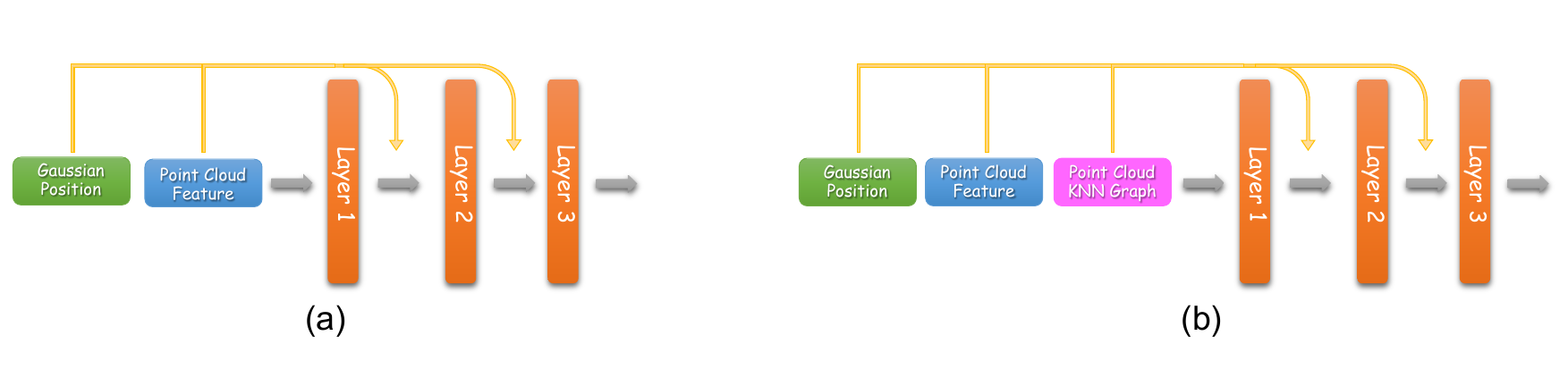}
    \caption{The architecture of MLPs. (a). The architecture of Spherical Harmonics MLP. (b). The architecture of Scale, Rotation and Opacity MLPs. }
    \vspace{-3mm}
    \label{mlparchitecture}
\end{figure*}

\subsection{Architecture of Point Transformer} \label{ptss}

We present the architectures of the point transformers, denoted as  \(\Theta_1\) and \(\Theta_2\), in Fig. \ref{pointransformer1} and Fig. \ref{pointransformer2}, respectively. The architecture of the MLPs is depicted in Fig. \ref{mlparchitecture}. Initially, the human point cloud is fed into a position encoding module to enable the network to learn high-frequency features. Subsequently, a point transformer is employed to extract point-wise features from the human point cloud. These point-wise features are then concatenated with the human point cloud and input into various MLPs to learn different 3D Gaussian attributes. In the second stage of overfitting, we also incorporate spherical harmonics features into the point transformer to unify the distribution across different scenes. For the MLPs responsible for Scale, Rotation, and Opacity, we further enhance their geometric perception by inputting the KNN graph of the point cloud.

\begin{figure*}[t!]
    \centering
    \setlength{\abovecaptionskip}{0.1cm}
    \includegraphics[width=5.4in]{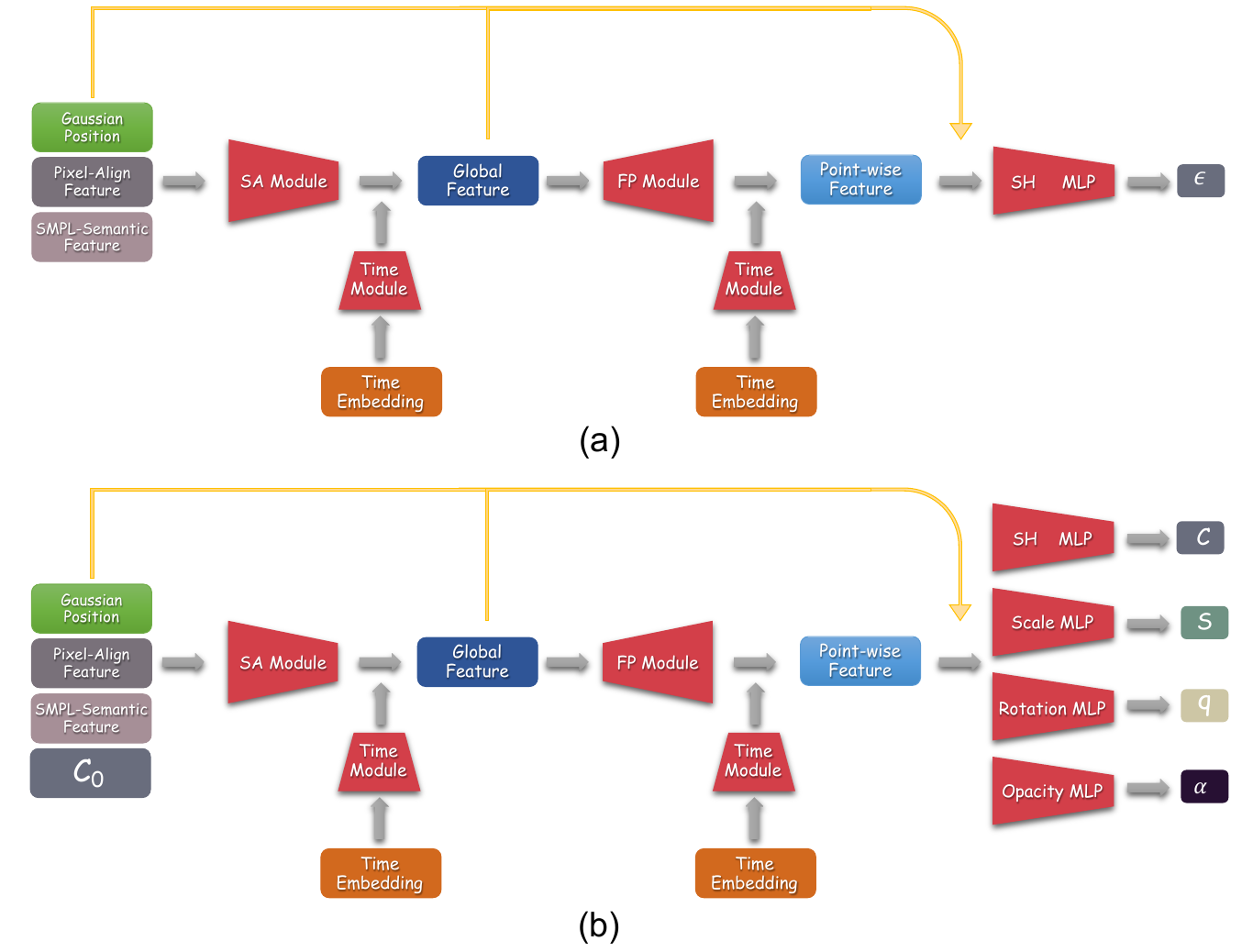}
    \caption{(a). The architecture of the 3D Gaussian attribute diffusion model. (b). The architecture of the 3D Gaussian attribute diffusion model to train the extra step. }
    \vspace{-3mm}
    \label{diffuserarchitecture}
\end{figure*}

\subsection{Architecture of 3DGS Diffuser} \label{3ddffs}

Currently, most point cloud-based diffusion models are designed for point cloud generation, yet they lack the capability to directly apply diffusion on 3D Gaussian attributes. To address this, we adopt PointNet++ as the backbone and introduce modifications to enable the training of a diffusion model. The architecture of the diffuser is illustrated in Fig. \ref{diffuserarchitecture}.

\begin{figure*}[t!]
    \centering
    \includegraphics[width=6.30in]{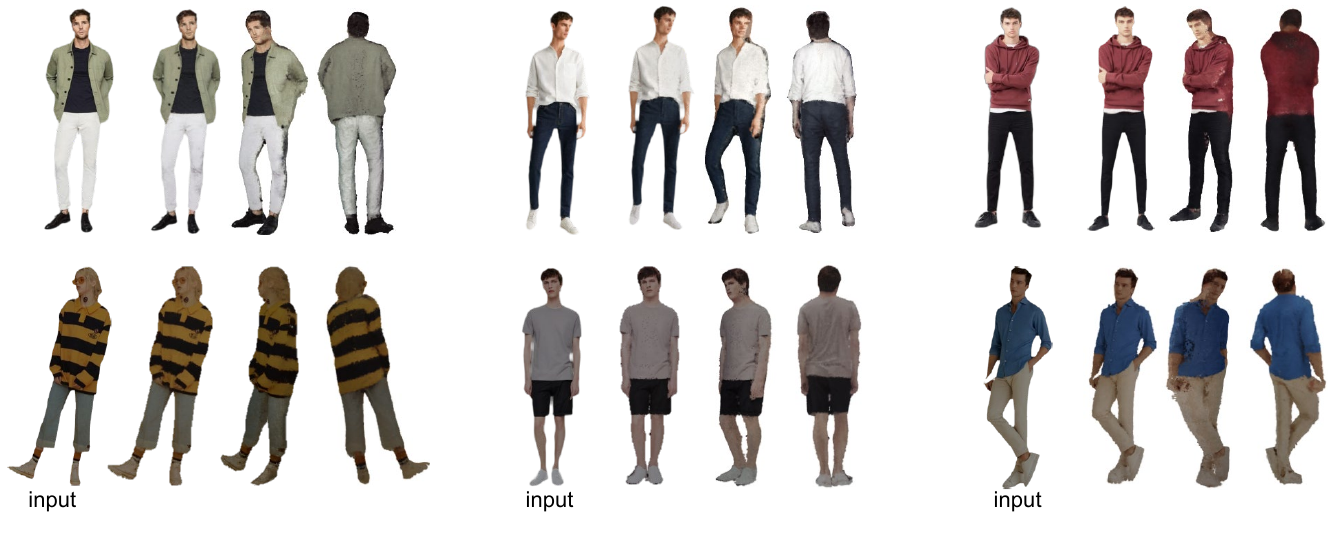}
    \caption{Visual results on in-the-wild images.}
    \label{r2q4fig}
\end{figure*}

\section{Additional Results on In-the-wild Images} We provide additional visual results on in-the-wild images in Fig. \ref{r2q4fig} to better demonstrate the generalizability of HuGDiffusion.
\vspace{2cm}

\end{document}